\documentclass[10pt,letterpaper]{article}

\usepackage{paper}      
\usepackage{logo}      

\newcommand{\ours}{X-Lens\xspace}
\newcommand{\omniscene}{OmniScene}
\newcommand{\omniscenesingle}{\omniscene-Single}
\newcommand{\omniscenequad}{\omniscene-Quad}
\newcommand{\omniscenefull}{\omniscene-Full}

\usepackage{bm}
\usepackage{multirow}

\definecolor{cvprblue}{HTML}{f16538}
\usepackage[pagebackref,breaklinks,colorlinks,allcolors=cvprblue]{hyperref}
\usepackage[most]{tcolorbox}
\usepackage{ragged2e}

\title{\ours: Real-Time Metric Depth Estimation with Heterogeneous Cameras}

\newcommand{\paperauthors}{%
{\bfseries Heng Zhou$^{1,*}$ \quad Shuhong Liu$^{1,2,*}$ \quad Yonghao He$^{1}$ \quad Bohao Zhang$^{1}$ \quad Fa Fu$^{1}$ \quad Chenhui Hou$^{1}$ \quad Xianbao Hou$^{1,3}$}\\
{\bfseries Lijun Han$^{1}$ \quad Cong Yang$^{3}$ \quad Wei Sui$^{1,\dagger}$}\\[0.5em]
\small $^{1}$D-Robotics \quad $^{2}$The University of Tokyo \quad $^{3}$Soochow University\\[0.2em]
\small $^{*}$Equal Contribution \quad $^{\dagger}$Project Lead%
}

\begin{document}

\definecolor{D-color}{RGB}{255,60,0}
\definecolor{CoverOrange}{RGB}{255,246,238}

\vspace*{-1.8em}

\begin{center}
\begin{tcolorbox}[
  enhanced,
  width=\textwidth,
  boxrule=0pt,
  frame hidden,
  arc=10pt,
  left=0.42cm,
  right=0.42cm,
  top=0.38cm,
  bottom=0.35cm,
  colback=CoverOrange,
  opacityback=0.8,
  before skip=0pt,
  after skip=0pt,
  overlay={\node[anchor=south east,xshift=-0.42cm,yshift=0.34cm] at (frame.south east) {\includegraphics[width=2.2cm]{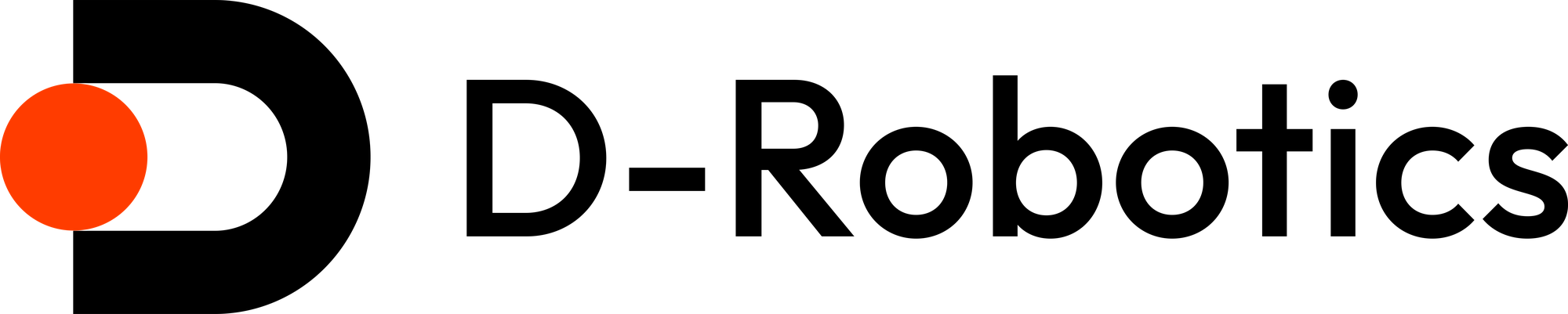}};}
]
\setlength{\parindent}{0pt}
\setlength{\parskip}{0.48em}
\raggedright
{\huge\bfseries \thetitle\par}
\vspace{0.35em}
{\small
\lineskip .25em
\paperauthors\par}
\vspace{0.75em}
{\normalsize
\begingroup
\justifying
\renewenvironment{abstract}{\par}{\par}
\begin{abstract}
We present \ours, a compact feed-forward model for metric depth estimation from a variable number of calibrated fisheye and pinhole views. To support real-time downstream perception, \ours is built around a geometry-aware heterogeneous camera formulation with two key components. Learnable calibration tokens provide a coarse alignment between fisheye and pinhole projective spaces, while a Jacobian-parameterized distortion bias injected into cross-attention models local projection changes and promotes cross-camera consistency, enabling robust generalization with only 0.04B parameters and up to 41 FPS. The model predicts dense depth together with a global metric scale, avoiding auxiliary reconstruction targets that increase computation and optimization complexity. To learn such cross-camera generalization at scale and depth, \ours is trained on multiple public datasets and OmniScene, our newly released large-scale synthetic dataset containing approximately 266K synchronized six-view frames, 1.7M individual images, and 103 indoor and outdoor scenes. Extensive experiments on both real-world and synthetic indoor and outdoor datasets demonstrate superior heterogeneous-camera metric depth accuracy, reducing AbsRel by 25.4\% on OmniScene-Full over the strongest baseline while using 88.9\% fewer parameters, with competitive performance on conventional fisheye-only and pinhole-only settings.
\end{abstract}
\endgroup
}
\vspace{0.2em}
{\small
\textbf{Correspondence:} \texttt{wei.sui@d-robotics.cc} \\
\textbf{Project Page:} \url{https://drobotics-xlens.github.io} \\
\textbf{Code:} \url{https://github.com/zhouhengamerica/XLens}
}
\end{tcolorbox}
\end{center}

\vspace{-2em}

\begin{center}
\centering
\includegraphics[width=.99\textwidth]{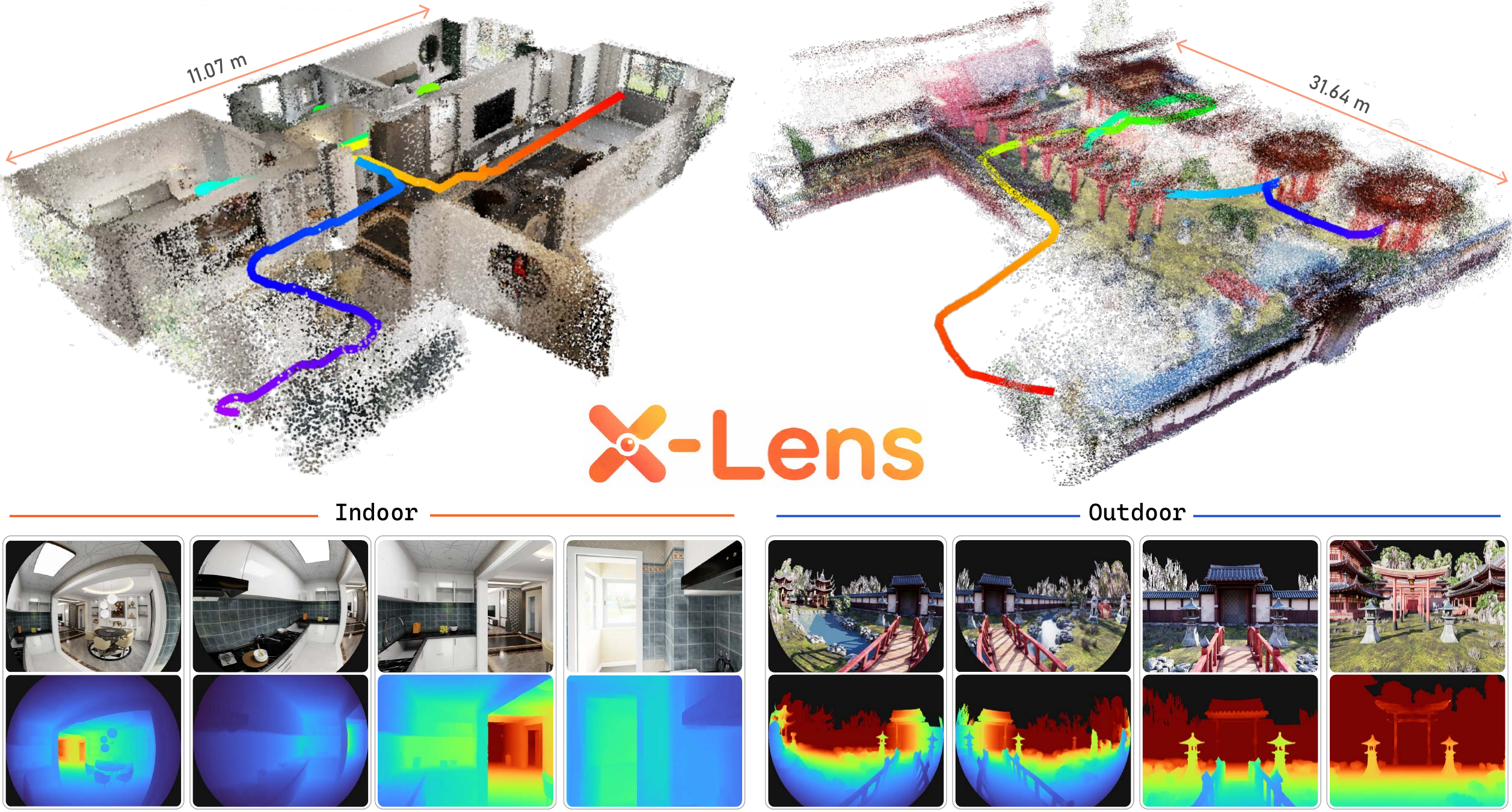}
\vspace{-0.7em}
\captionof{figure}{Teaser illustration of \ours. The top row shows metric point clouds projected from calibrated six-view heterogeneous cameras for two scenes, demonstrating metric depth estimation across mixed camera rigs. The bottom row shows RGB inputs and predicted depth maps from heterogeneous multi-view pinhole and fisheye cameras.}
\label{fig:teaser_cover}
\end{center}

\vspace{0.6em}

\section{Introduction}
\label{sec:intro}

The ability to perceive and interpret 3D spatial information from visual inputs is a cornerstone of spatial intelligence~\cite{zhen20243dvla,zheng2024occworld}, providing the geometric foundation for intelligent systems to build persistent world representations~\cite{zheng2024occworld,bruce2024genie}, reason about physical scenes~\cite{wang2023dust3r,wang2025vggt}, and act safely in open-ended environments~\cite{wang2023voyager,brohan2023rt2}. As autonomous robots~\cite{brohan2023rt2,team2024octo}, vision-language-action (VLA) models~\cite{brohan2023rt2,kim2024openvla,zhen20243dvla}, and physical world models~\cite{zheng2024occworld,bruce2024genie} are increasingly deployed for fine-grained spatial reasoning~\cite{zhen20243dvla} and long-horizon execution~\cite{wang2023voyager,team2024octo}, they impose stricter geometric requirements on fundamental spatial perception to support planning~\cite{brohan2023rt2}, interaction~\cite{kim2024openvla}, generalization~\cite{zhen20243dvla}, and behavioral robustness~\cite{zheng2024occworld}.

In modern embodied systems, these requirements are commonly supported by multi-camera 3D perception. Multiple views expand spatial coverage, reduce occlusion, and provide complementary geometric cues for scene understanding. Robot policies commonly combine workspace and wrist-mounted cameras to capture both task-level context and local manipulation details~\cite{brohan2023rt2,kim2024openvla,team2024octo}, while autonomous driving and navigation systems use surround-view rigs for wide-area perception, occupancy reasoning, and planning~\cite{zheng2024occworld,wang2023voyager}. Such systems often employ cameras with different viewpoints, resolutions, and fields of view. Narrow- or standard-FOV pinhole cameras preserve distant details, whereas wide-FOV or fisheye cameras provide near-field and peripheral coverage.

However, existing 3D and depth foundation models face substantial limitations in this heterogeneous-camera setting. Current depth estimation methods~\cite{bhat2023zoedepth,yin2023metric3d,piccinelli2024unidepth,yang2024depthanythingv2} and multi-view geometric methods~\cite{wang2023dust3r,leroy2024mast3r,wang2025vggt,wang2025pi3,keetha2025mapanything,wang2026vggtomega} are typically built around a single camera model, pinhole-centric assumptions, or reconstruction objectives that are difficult to satisfy under real-time constraints. In mixed fisheye-pinhole rigs, the non-uniform projection geometry and view-dependent distortion introduce heterogeneous ray distributions, making cross-view feature correspondence and metric-scale alignment difficult. Panorama-based formulations~\cite{jiang2021unifuse,shen2022panoformer,ai2025da360,lin2025depthanypanoramas} simplify omnidirectional processing, but they may alter the native image geometry and underuse co-visible cues across calibrated views. Existing geometric foundation models~\cite{wang2023dust3r,leroy2024mast3r,wang2025vggt,wang2025pi3,keetha2025mapanything,wang2026vggtomega} are also computationally demanding, making edge-side inference difficult under the latency and memory budgets of real-time downstream systems. These trade-offs motivate a compact formulation that focuses on depth and scale while natively supporting heterogeneous multi-view inputs.


To address these challenges, we present \textbf{\ours}, a feed-forward model for metric depth estimation from calibrated heterogeneous camera rigs, with representative predictions shown in \Cref{fig:teaser_cover}. Given arbitrary fisheye and pinhole views in their native image domains, \ours predicts dense depth maps and a global scale factor through a compact DINOv2~\cite{oquab2024dinov2} backbone, DPT~\cite{ranftl2021dpt} depth head, and scale-regression MLP. Rather than depending on large-scale pre-trained backbones~\cite{oquab2024dinov2,Simeoni2025DINOv3}, the model learns transferable depth representations through staged training on calibrated heterogeneous multi-view data. The training process first establishes robust pinhole depth priors, then adapts them to wide-FOV observations with learnable calibration tokens. To further bridge camera-specific projection geometries, we inject a distortion bias derived from patch-center Jacobians and local ray directions into global cross-attention, allowing attention weights to account for local projection changes across views. This design improves cross-camera feature consistency while preserving each camera's native sampling pattern, avoiding panorama conversion, and eliminating auxiliary reconstruction targets that are unnecessary for metric depth inference.

\begin{figure}[t]
\centering
\includegraphics[width=0.95\textwidth]{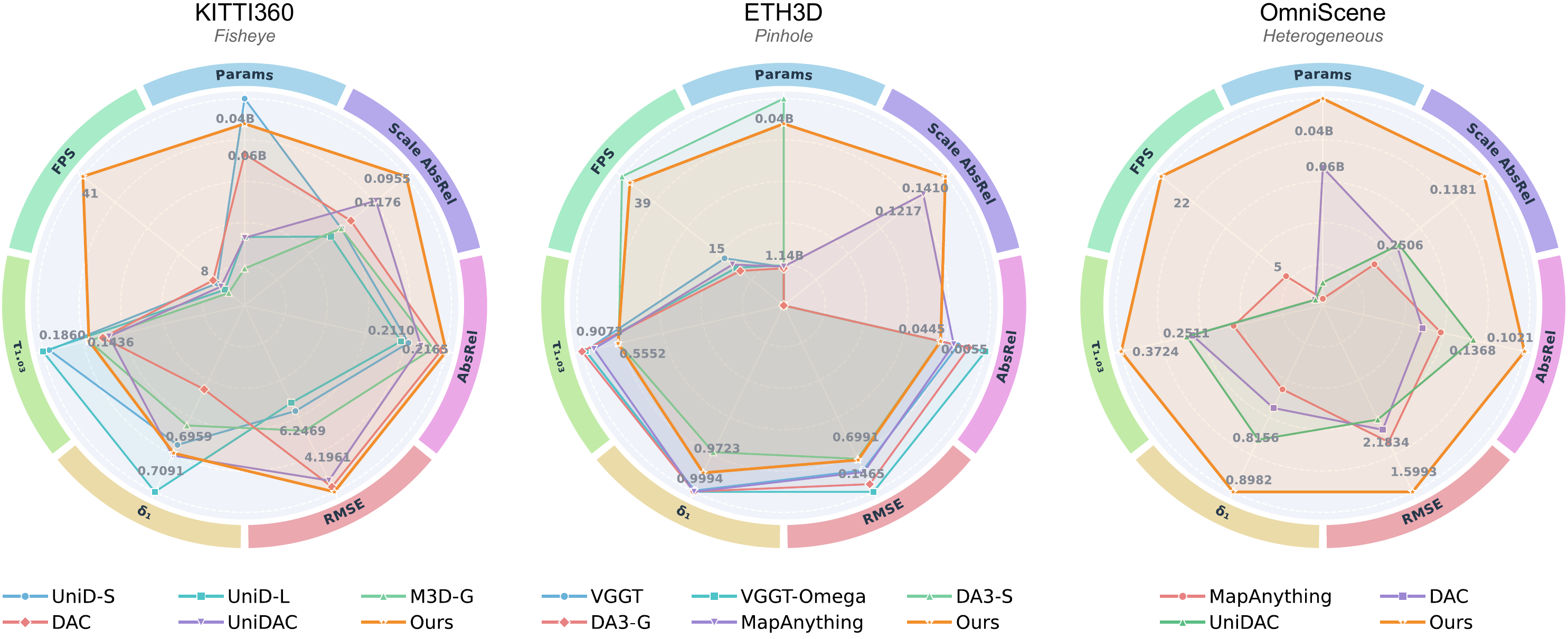}
\vspace{-0.6em}
\caption{Radar-chart comparison across fisheye, pinhole, and heterogeneous-camera benchmarks. \ours consistently delivers strong accuracy and efficiency across all camera settings.}
\label{fig:radar_map}
\vspace{-0.4em}
\end{figure}

Furthermore, to overcome the scarcity of heterogeneous-camera depth training corpora, we construct a large-scale synthetic dataset with 266,000 frames across 103 indoor and outdoor scenarios. It provides diverse scene layouts, high-quality rendering, and calibrated camera configurations for omnidirectional and heterogeneous depth perception. By combining this synthetic corpus with existing training data, \ours learns robust metric depth priors and demonstrates strong zero-shot generalization on unseen environments. With a single feed-forward pass, it produces accurate metric depth estimates for complex scenes, making it suitable for real-time downstream spatial intelligence applications. \Cref{fig:radar_map} summarizes this balance of accuracy and efficiency across fisheye, pinhole, and heterogeneous-camera settings.

In summary, our main contributions are as follows:\vspace{0.6em}
{\setlength{\parskip}{0pt}
\begin{itemize}[leftmargin=*,labelsep=0.4em]
\item \textbf{Efficient Heterogeneous Camera Depth Estimation Model:} We propose \ours, a lightweight 0.04B-parameter feed-forward metric depth estimation model that natively supports raw multi-view heterogeneous inputs without panoramic stitching.
\item \textbf{Large-scale Dataset:} We curate a high-quality heterogeneous dataset with 266,000 frames across 103 scenarios, providing a comprehensive resource for depth perception research with improved diversity and realism.
\item \textbf{Superior Zero-shot Generalization and Real-Time Performance:} Extensive experiments show that \ours achieves state-of-the-art metric depth accuracy with strong zero-shot generalization. Its lightweight design enables high-frame-rate inference with low latency, bridging metric depth modeling and real-time downstream deployment.
\end{itemize}
}

\section{Related Works}
\label{sec:formatting}

\paragraph{Monocular Depth Estimation.}
Monocular depth estimation has progressed from dataset-specific regression toward generalizable depth foundation models. Early supervised CNN and Transformer methods~\cite{eigen2014depth,lee2019bts,bhat2021adabins,yuan2022newcrfs} improved metric depth accuracy on standard indoor and outdoor benchmarks, while self-supervised methods~\cite{godard2019monodepth2,guizilini2020packnet} reduced the need for dense depth labels through photometric consistency. Subsequent works shifted toward cross-dataset generalization, with MiDaS~\cite{ranftl2020midas} and DPT~\cite{ranftl2021dpt} learning affine-invariant relative depth from mixed datasets, and ZoeDepth~\cite{bhat2023zoedepth} combining relative pretraining with metric heads for zero-shot metric transfer. More recent foundation models further scale data, supervision, and backbone capacity, with DepthAnything~\cite{yang2024depthanything,yang2024depthanythingv2} emphasizing large-scale unlabeled or pseudo-labeled training for robust relative depth, Metric3D~\cite{yin2023metric3d,hu2024metric3dv2} learning canonical-camera metric depth, UniDepth~\cite{piccinelli2024unidepth,piccinelli2025unidepthv2} estimating universal metric depth with camera-aware representations, Depth Pro~\cite{bochkovskii2024depthpro} producing sharp zero-shot metric depth and focal length, and MoGe~\cite{wang2024moge,wang2025moge2} recovering monocular 3D geometry with stronger metric scale and fine details. Recent monocular depth studies also continue to improve high-resolution refinement~\cite{li2025patchrefinerv2}, temporal consistency for video~\cite{yang2025videodepthanything}, generative dense prediction priors~\cite{he2024lotus}, camera-agnostic 3D representations~\cite{piccinelli2025unik3d}, and metric grounding or scale recovery~\cite{wang2026metricanything,li2026languageprior,wang2026anchord}. These monocular methods focus on estimating depth from a single image, while \ours targets multi-view depth, where cross-camera constraints can be exploited jointly rather than inferred from one image alone.

\paragraph{Wide-FOV Depth Estimation.}
Wide-FOV depth estimation has developed along two main directions, panoramic depth estimation and direct wide-angle processing. Panorama-based methods project inputs into an Equirectangular Projection and estimate depth on the spherical image domain. Early dual-projection methods~\cite{wang2020bifuse,jiang2021unifuse} fused equirectangular and cubemap features to reduce polar distortion, while HoHoNet~\cite{sun2021hohonet}, PanoFormer~\cite{shen2022panoformer}, and EGformer~\cite{yun2023egformer} introduced layout cues, tangent-plane partitioning, and spherical geometry biases for stronger panoramic depth estimation. Recent panoramic foundation models further adapt large monocular depth backbones to wide-FOV inputs, with DepthAnything in $360^{\circ}$~\cite{ai2025da360}, DepthAnything in Any Direction~\cite{xu2025da2}, DepthAnyCamera~\cite{zhang2025depthanycamera}, UniDAC~\cite{ganesan2026unidac}, and DepthAnyPanoramas~\cite{lin2025depthanypanoramas} improving scale-invariant, camera-agnostic, and zero-shot metric panoramic depth. Direct wide-angle methods instead avoid full panorama conversion and estimate depth in fisheye or multi-fisheye image spaces, including distortion-aware monocular depth~\cite{tateno2018distortion,li2021fisheyedistance}, spherical cost-volume stereo~\cite{won2020omnimgs,meuleman2021spheresweeping}, and learning-based multi-fisheye systems such as OmniMVS~\cite{wang2022omnimvs}, Unsupervised OmniMVS~\cite{wang2023unsupomnimvs}, and CasOmniMVS~\cite{li2024casomnimvs}. Recent wide-FOV studies also continue to improve fisheye benchmarks~\cite{lu2026widedepth}, camera-model conditioning~\cite{jung2026wid3r}, and generalization across unseen camera types~\cite{zhang2025depthanycamera,lin2025depthanypanoramas}. These methods mainly rely on panorama resampling, single-view wide-FOV inference, or hardware-specific stereo volumes, while \ours keeps the input in its native camera views and learns cross-view depth fusion without converting the observations into a stitched panorama.

\paragraph{Feed-forward Geometric Models.}
Recent feed-forward geometry foundation models provide a new route for multi-view depth estimation by learning dense 3D structure directly from image collections. DUSt3R~\cite{wang2023dust3r} introduced pointmap regression in a shared coordinate frame for dense matching, pose recovery, and depth reasoning, and MASt3R~\cite{leroy2024mast3r} further grounded image matching in this 3D representation. Subsequent works extend this feed-forward paradigm across view count and scene dynamics, with Spann3R~\cite{wang2024spann3r} accumulating geometry with spatial memory, MonST3R~\cite{zhang2024monst3r} handling dynamic scenes, and Fast3R~\cite{yang2025fast3r} targeting large image collections. VGGT~\cite{wang2025vggt} further unified depth, point maps, intrinsics, and poses in a single visual geometry Transformer. $\pi^3$~\cite{wang2025pi3} then studied scalable permutation-equivariant visual geometry learning. MapAnything~\cite{keetha2025mapanything} extended this direction toward universal feed-forward metric 3D reconstruction. DepthAnything3~\cite{lin2025depthanything3} further extends monocular depth priors to any-view visual-space recovery. Recent feed-forward multi-view depth and reconstruction studies also continue to improve generality under sparse-view~\cite{chen2026reliev3r,lin2026mix3r,burzio2026dejaview}, zero-shot MVS~\cite{izquierdo2025mvsanywhere}, synthetic-data~\cite{ma2026simpleproc}, long sequences~\cite{jin2026zipmap,xie2026scal3r} and large training corpora~\cite{wang2026vggtomega}, as well as high-fidelity reconstruction~\cite{jiang2025anysplat,zhao2026structsplat} and wide-FOV camera models~\cite{jung2026wid3r}. These works provide strong priors for multi-view geometry, but most are still optimized for general reconstruction where depth, pose, intrinsics, and point maps are inferred jointly. For heterogeneous multi-view rigs, such coupling and the common pinhole-centric design make it difficult to robustly support fisheye cameras and mixed sensor configurations. \ours instead focuses on heterogeneous multi-view metric depth with a compact perception-oriented design, making real-time inference and edge deployment practical.
\section{Method}
\label{sec:method}

\begin{figure}[!tp]
\centering
\includegraphics[width=\textwidth]{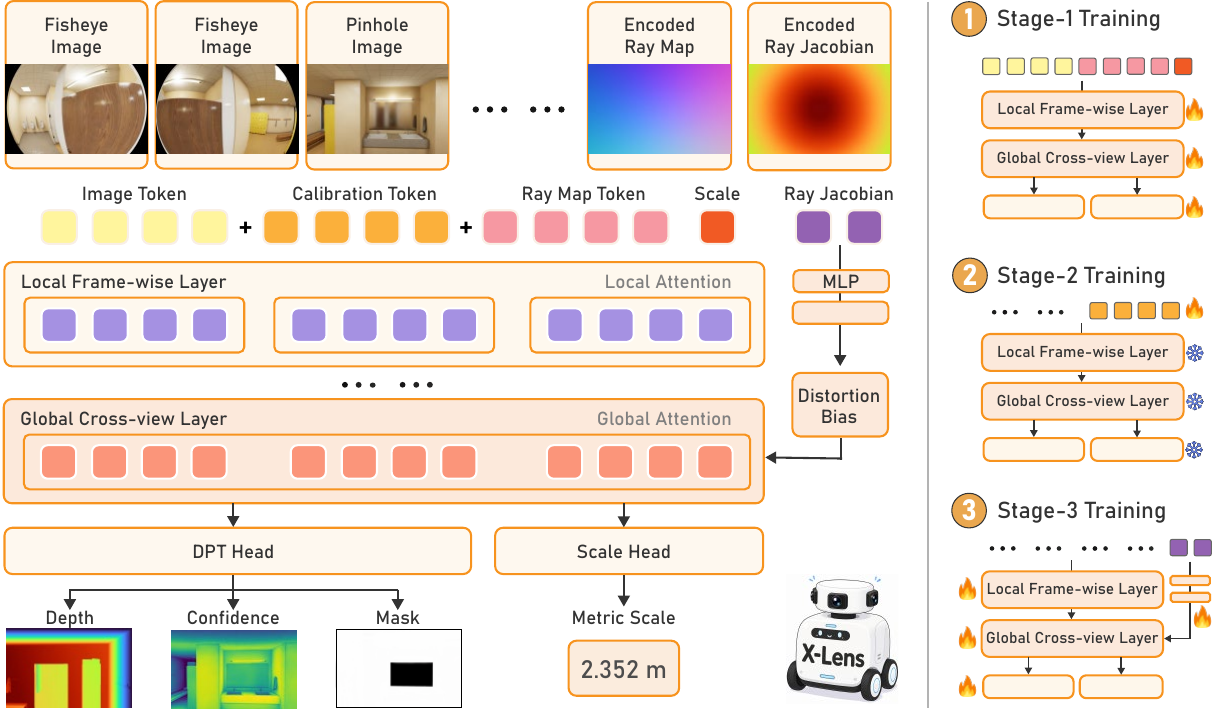}
\caption{Overview of the proposed \ours pipeline. Heterogeneous multi-view inputs are represented through generic camera rays and processed by an omnidirectional heterogeneous projection transformer with calibration tokens and distortion-aware attention, followed by factored prediction heads for normalized depth, confidence, and global metric scale.}
\label{fig:pipeline}
\end{figure}

We present \textbf{\ours}, a feed-forward multi-view depth estimation model that predicts an up-to-scale depth field, an associated confidence map, and one global metric scalar. We first formulate heterogeneous cameras with a generic ray representation and define the factored depth output in \S\ref{sec:method:formulation}. We then describe the heterogeneous projection transformer that combines calibration tokens with distortion-aware cross-view attention in \S\ref{sec:method:trunk}. Next, we introduce the geometric position embedding and scale attention mechanism used for robust metric scale prediction in \S\ref{sec:method:grope_scale}. Finally, we present the three stage training pipeline and optimization objective in \S\ref{sec:method:training_pipeline}. \Cref{fig:pipeline} provides an overview of the proposed pipeline and training strategy.

\subsection{Formulation and Design Principles}
\label{sec:method:formulation}

\paragraph{Generic Heterogeneous Camera Representation}
\label{sec:method:generic_cam}

Existing 3D depth and foundation models are commonly centered on the pinhole camera model. They ingest a linear intrinsic matrix $\mathbf{K} \in \mathbb{R}^{3 \times 3}$ and assume a one-to-one correspondence between image coordinates and 3D rays via $\mathbf{r} \propto \mathbf{K}^{-1} [u, v, 1]^\top$. However, this assumption fails under strongly non-linear lenses such as fisheye, where the projection is governed by a parametric distortion 
vector rather than a few scalars. To handle multi-view heterogeneous configurations, 
\ours abstracts cameras away from specific projection models and replaces 
the pinhole intrinsic with a generic unprojection map $\mathcal{G}$. Each 
view $s$ is formally specified as:
\begin{equation}
\mathbf{r}_{s,p} = \mathcal{G}(p; \bm\xi_s, \tau_s) \mathbf{R}_s,
\qquad
\mathbf{r}_{s,p} \in \mathbb{S}^{2}, \; p \in [0,W] \times [0,H],
\label{eq:generic_unproject}
\end{equation}
where $\mathcal{G}$ maps a continuous pixel coordinate $p$ to a 
back-projected unit ray. It absorbs both the per-camera calibration 
$\bm\xi_s$, including focal lengths, principal point, and distortion coefficients, and an explicit camera-type indicator $\tau_s$, with $\tau_s \in \{\text{pinhole}, \text{fisheye}\}$. $\mathbf{R}_s \in SO(3)$ denotes 
the rotation matrix in the rig frame.

Crucially, all downstream geometric reasoning inside the network is 
conducted in the ray space $\{\mathbf{r}_{s,p}\}$ rather than the pixel coordinate space. For cameras without a closed-form projection model, we 
implement $\mathcal{G}$ using a tabulated radial profile sampled during 
training. Consequently, $\mathcal{G}$ can be flexibly specified per dataset 
and per sensor modality. Without any architectural modifications, the same 
network can directly ingest heterogeneous mixtures of pinhole, fisheye, 
and $360^{\circ}$ cameras.

\paragraph{Factored Depth Prediction}
\label{sec:method:outputs}

Given multiple heterogeneous views $\mathcal{I} = \{I_s \in \mathbb{R}^{3 \times 
H \times W}\}_{s=1}^{S}$ and their corresponding calibration parameters 
$\{\bm\xi_s, \mathbf{T}_s\}_{s=1}^{S}$, \ours predicts a factored 
triplet:
\begin{equation}
\Big[ \hat D, \; \hat C, \; \hat m \Big],
\quad
\hat D \in \mathbb{R}_{>0}^{S \times H \times W}, \;
\hat C \in \mathbb{R}_{>1}^{S \times H \times W}, \;
\hat m \in \mathbb{R}_{>0},
\label{eq:outputs}
\end{equation}
where $\hat D$ is the normalized depth field, defined as the per-pixel $z$-depth (depth along the camera optical axis, \emph{not} the Euclidean ray distance) divided by the in-batch mean to remove absolute scale, $\hat C$ is the self-calibrated confidence map, and $\hat m$ is a single global scalar. This $z$-depth convention is kept consistent with the ground-truth used in our evaluation, where Euclidean fisheye depth is converted to $z$-depth. The final metric depth is then reconstructed via a single multiplication: $\hat D^{\text{m}} = \hat m \cdot \hat D$. Besides this factored triplet, the shared DPT head that regresses $\hat D$ and $\hat C$ also produces, as an additional output channel, a binary validity mask $\hat M \in [0,1]^{S \times H \times W}$ that flags invalid regions such as sky or pixels outside the physical lens mask; it only gates the training loss \cref{eq:total_loss_expanded} and is not part of the metric output.

Compared with recent any-view foundation models that typically predict numerous geometric factors such as ray maps, depth scales, and camera poses~\cite{wang2023dust3r,wang2025vggt,keetha2025mapanything}, our three-tuple design is intentionally streamlined for omnidirectional perception scenarios where rig calibration is usually known during deployment and the ultimate target is metric depth. We concentrate all calibration and pose priors on the input side while keeping the output side focused on depth, uncertainty, and global scale. Crucially, this design isolates the absolute physical scale into a one-dimensional pathway through $\hat m$, allowing all other geometric losses to operate in a scale-invariant normalized space. By expressing all geometric variables relative to the canonical reference camera frame, the network can absorb heterogeneous resolution scales through intrinsic-guided normalization while avoiding dependence on an arbitrary world coordinate system during inference.

\subsection{Heterogeneous Projection Transformer}
\label{sec:method:trunk}
The core of \ours is built upon a Vision Transformer~\cite{dosovitskiy2021vit} backbone with alternating within-view layers $L_s$ and cross-view layers $L_g$ for local feature extraction and global context aggregation. Since standard multi-view attention does not model the heterogeneous imaging topologies of pinhole and fisheye cameras, we introduce \textbf{\textit{Multi-View Calibration Tokens}} to absorb lens-specific distortion, inspired by calibration-token adaptation~\cite{gangopadhyay2025calibtokens}, and \textbf{\textit{Jacobian Distortion Bias}} to inject cross-lens geometric correspondence priors into global attention. Together with FishRoPE~\cite{ahuja2026fishrope} and the factored prediction head, these modules form a compact architecture for heterogeneous metric reconstruction.

\paragraph{Multi-View Calibration Tokens}
\label{sec:method:calib_tokens}

Heterogeneous inputs carry distinct per-lens distortion patterns. To keep these from being absorbed indiscriminately by the shared visual tokens, we introduce a learnable calibration-token tensor $\mathbf{\Theta} \in \mathbb{R}^{N_L \times T \times K \times C}$ indexed by Transformer layer $N_L$, camera type $T$, token count $K$, and dimension $C$. At each layer $i$, the type-specific slice $\mathbf{\Theta}[i, \tau_s]$ is appended to view $s$'s token sequence, attends, and is then dropped before the next layer re-injects a fresh slice, forming an inject-attend-drop process. A slice is injected at every layer, including the cross-view layers. Within a cross-view layer, an attention mask keeps each token view-local, so it attends only to its own view and adds no camera-type signal to the global metric fusion. The drop serves a separate purpose: it prevents the tokens from persisting across layers. Tokens are injected only for fisheye views, leaving the pinhole pathway unchanged, and $\mathbf{\Theta}$ is zero-initialized so that corrections are learned gradually from an identity start.

This design differs from the monocular calibration tokens of~\cite{gangopadhyay2025calibtokens}, where a single token set is inserted once at the input and aligns fisheye features to the perspective latent space through self- and cross-attention. Our tokens are instead layer- and camera-type-specific, and although they enter the cross-view layers, the attention mask keeps them view-local. This confines lens correction to each individual view and protects the cross-view metric fusion that our heterogeneous setting relies on.

\paragraph{Jacobian Distortion Bias}
\label{sec:method:distortion_bias}

Calibration tokens provide view-local adaptation for lens-specific distortions, whereas reliable heterogeneous matching also requires cross-view geometric compatibility. This is particularly important for fisheye--pinhole cross-view interactions, where visually corresponding regions can occupy very different image-plane neighborhoods due to non-linear projection. We therefore introduce the Jacobian Distortion Bias in the cross-view attention layers $L_g$. The bias is applied to all patch tokens and acts as a geometry-aware prior that modulates attention according to local ray orientation and projection distortion.

The bias is computed from the ray field already available to the network. We reuse the per-pixel ray field $\{\mathbf{r}_{s,p}\}$ from \cref{eq:generic_unproject} and downsample it to the ViT patch grid to obtain a per-patch ray $\mathbf{r} \in \mathbb{S}^2$, where $\mathbf{r}_i$ denotes the ray at patch $i$. Local projection behavior is characterized by the unprojection Jacobian $\mathbf{J} \in \mathbb{R}^{3 \times 2} = [\partial\mathbf{r}/\partial u, \partial\mathbf{r}/\partial v]$, estimated by finite differences on the patch grid. The scalar $s = \|\mathbf{J}\|_F$ captures local spatial expansion, while the full Jacobian retains anisotropic stretching and orientation cues that are important near strongly distorted fisheye regions.

For patch tokens indexed by $i$ and $j$, we form a relative descriptor 
$\bm{\phi}_{ij} \in \mathbb{R}^9$:
\begin{equation}
\bm{\phi}_{ij} = \Big[ \mathbf{r}_i \cdot \mathbf{r}_j, \;
\log\|\mathbf{J}_i\|_F - \log\|\mathbf{J}_j\|_F, \;
(\mathbf{r}_i - \mathbf{r}_j)^\top, \;
\text{vec}(\mathbf{J}_i^\top \mathbf{J}_j)^\top \Big]^\top,
\label{eq:distortion_descriptor_en}
\end{equation}
which combines angular consistency, relative local scale, ray displacement, and Jacobian correlation. A lightweight head-specific MLP converts this descriptor into an additive bias for the $k$-th attention head:
\begin{equation}
\text{Attention}^{(k)}(\mathbf{Q}, \mathbf{K}, \mathbf{V}) = 
\text{Softmax}\left( \frac{\mathbf{Q}^{(k)} (\mathbf{K}^{(k)})^\top}{\sqrt{d_k}} 
+ \mathbf{B}^{(k)} \right) \mathbf{V}^{(k)},
\label{eq:attention_bias_formula_en}
\end{equation}
\begin{equation}
\mathbf{B}_{ij}^{(k)} = \text{MLP}^{(k)}(\bm{\phi}_{ij}),
\label{eq:bias_mlp_formula_en}
\end{equation}
where $\mathbf{B}^{(k)} $ is added before Softmax normalization. This formulation encourages cross-view attention to favor patches with compatible 3D ray geometry and local projection structure, rather than relying solely on appearance similarity in distorted image coordinates.

\paragraph{Geometric Rotary Position Embedding and Scale Attention}
\label{sec:method:grope_scale}

To make positional encoding consistent with the ray-space representation used by heterogeneous cameras, we build our geometric rotary embedding on FishRoPE~\cite{ahuja2026fishrope}. Using the same per-patch ray $\mathbf{r} \in \mathbb{S}^2$ defined above, the embedding parameterizes rotary phases by relative ray orientation and modulates the Query--Key interaction as $\mathbf{q}_i^\top \mathbf{R}_{\mathbf{r}_j - \mathbf{r}_i} \mathbf{k}_j$~\cite{ahuja2026fishrope}. Since $\mathbf{r}$ is derived from the generic unprojection map $\mathcal{G}$ for both pinhole and fisheye views, the same embedding form operates across different projection models.


While the ray aware embedding improves geometric feature alignment, metric scale regression is still sensitive to the spatial reliability of the pooled features. In fisheye views, peripheral regions are radially compressed and tend to carry larger geometric uncertainty. Directly pooling all spatial features to regress the metric scale $\hat{m}$ can therefore bias the prediction toward distorted regions with low reliability.

For robust metric scale estimation, we introduce Scale Attention, a confidence guided pooling mechanism that uses the predicted confidence map $\hat{C}$ to select spatially reliable regions. The lowest confidence $25\%$ of pixels are discarded, and confidence weighted pooling is applied over the remaining core set $\Omega_{\text{core}}$ as:
\begin{equation}
\hat{m} = \text{MLP}\left( \frac{\sum_{p \in \Omega_{\text{core}}} \hat{C}_p \cdot \mathbf{F}_p}{\sum_{p \in \Omega_{\text{core}}} \hat{C}_p} \right),
\label{eq:scale_attention_formula_en}
\end{equation}
where $\mathbf{F}_p$ denotes the spatial feature at pixel $p$. This restricts metric scale prediction to geometrically reliable regions and reduces sensitivity to fisheye boundary artifacts.

\subsection{Multi-Stage Heterogeneous Training Strategy}
\label{sec:method:training_pipeline}
\ours is trained with a progressive three-stage pipeline that moves from
homogeneous pinhole geometry to fisheye adaptation and finally to heterogeneous
multi-camera optimization. This schedule isolates lens-specific adaptation before
joint training, reducing interference between pinhole priors and fisheye distortion
modeling. The goal is to first obtain a stable multi-view geometric backbone,
then introduce fisheye-specific parameters under a controlled optimization regime,
and only afterward expose the full model to mixed-camera interactions.

\paragraph{Stage 1: Pinhole Pre-training.}
We first train the base network on multi-view pinhole data to learn scale-aware
geometric representations. The backbone and prediction heads are optimized while
calibration tokens and Jacobian distortion bias are disabled, so the model learns
standard perspective geometry without lens-specific adaptation. Since pinhole
datasets provide abundant and diverse multi-view supervision, this stage gives the
network a strong initialization for depth, confidence, and global scale estimation
before handling non-linear projection effects.

\paragraph{Stage 2: Fisheye Token Adaptation.}
We then adapt the model to omnidirectional fisheye inputs by freezing the backbone,
ray encoders, and prediction heads, and updating only the calibration tokens
$\mathbf{\Theta}$. This stage confines radial compression and non-linear lens
effects to the dedicated token pathway while preserving the pinhole representation
learned in Stage 1. By preventing the shared decoder and backbone from drifting,
the model learns lens-dependent corrections without overwriting the geometric
priors already established in the homogeneous domain.

\paragraph{Stage 3: Heterogeneous Joint Training.}
Finally, we activate the Jacobian Distortion Bias and jointly fine-tune the model
on mixed fisheye--pinhole multi-view data and the Stage-1 pure-pinhole datasets.
This stage trains cross-view attention
to reconcile heterogeneous image-plane distortions under a unified ray-space
representation. Since OmniScene's two pinhole views barely overlap, combining training with overlapping pure-pinhole
samples preserve the model's multi-view pinhole performance~\cref{sec:appendix_eval_detail}. Unlike Stage 2, which focuses on view-local fisheye adaptation,
this final stage optimizes the interactions between camera types, enabling the
model to match distorted fisheye regions with perspective pinhole observations in
a shared metric reconstruction framework.

\paragraph{Optimization Objectives.}
The entire optimization pipeline is driven by a factored multi-task objective 
function designed to explicitly decouple dense shape geometry from absolute 
physical scale. The total loss $\mathcal{L}$ is formulated as:
\begin{equation}
\mathcal{L} = \lambda_{\text{depth}} \mathcal{L}_D + \lambda_{\text{grad}} \sum_{k \in \{x,y\}} \Big\|\nabla_k \tfrac{\hat{D}}{\bar{\hat{D}}} - \nabla_k \tfrac{D}{\bar{D}}\Big\|_1 + \lambda_{\text{local}} \mathcal{L}_{\text{local}} + \lambda_{\text{scale}} \frac{|\hat{m} - m|}{m} + \lambda_{\text{mask}} \mathcal{L}_{\text{mask}},
\label{eq:total_loss_expanded}
\end{equation}
where $\Omega$ denotes the valid pixel set, $\nabla_x, \nabla_y$ represent
horizontal and vertical finite difference operators. Following $\mathcal{L}_D$, the gradient term is applied to the mean-normalized predicted and ground-truth depths $\hat{D}/\bar{\hat{D}}$ and $D/\bar{D}$ (with $\bar{\hat{D}}$ and $\bar{D}$ the in-batch means defined in \cref{eq:confidence_loss_final}), so that it remains scale-invariant and consistent with the normalized depth space rather than mixing normalized predictions with metric ground truth. The scaling factor $m$
provides the absolute physical metric anchor. 
The loss hyper-parameters are balanced via coefficients $\lambda_{\text{depth}}=1.0$, $\lambda_{\text{grad}}=1.0$, 
$\lambda_{\text{local}}=0.5$, $\lambda_{\text{scale}}=1.0$, and $\lambda_{\text{mask}}=0.2$. To eliminate 
absolute scale constraints during shape learning while dynamically modulating 
geometric uncertainty via the self-calibrated confidence map $\hat{C}$, the dense 
depth loss $\mathcal{L}_D$ is defined as:
\begin{equation}
\mathcal{L}_D = \frac{1}{|\Omega|} \sum_{p \in \Omega} \left( \hat{C}_p \cdot \left| \frac{\hat{D}_p}{\bar{\hat{D}}} - \frac{D_p}{\bar{D}} \right| - \lambda_{\text{conf}} \log \hat{C}_p \right),
\label{eq:confidence_loss_final}
\end{equation}
where $\bar{\hat{D}}$ and $\bar{D}$ denote the in-batch spatial means of the 
predicted and ground-truth depth maps, respectively, and $\lambda_{\text{conf}}=1.0$ 
is a regularization parameter preventing optimization collapse. The local term $\mathcal{L}_{\text{local}}$ enforces multi-scale local geometric consistency on the mean-normalized depth and follows the multi-scale local loss of MoGe~\cite{wang2024moge,wang2025moge2}. The mask term $\mathcal{L}_{\text{mask}}$ supervises the mask channel $\hat{M}$ of the shared DPT head against the ground-truth validity mask (sky or pixels outside the physical lens mask), following the mask supervision of DepthAnything3~\cite{lin2025depthanything3}.
\section{The OmniScene Dataset}
\label{sec:omniscene_dataset}

\paragraph{Overview}
To bridge the gap in multi-camera geometric perception, we introduce
\textbf{OmniScene}, a large-scale synthetic dataset for multi-view metric depth
estimation with heterogeneous cameras. It is rendered with a calibrated
six-camera rig containing four fisheye and two pinhole views. It targets two
key limitations in existing resources: (i) the scarcity of diverse wide-FOV
fisheye data, and (ii) the lack of synchronized heterogeneous benchmarks with
dense metric ground truth. Unlike driving-centric datasets limited
to outdoor road scenes, OmniScene covers diverse indoor and outdoor environments,
including residential spaces, commercial offices, shopping malls, industrial
warehouses, urban scenes, sci-fi complexes, and stylized period interiors. The
scenes are built from professionally authored Kujiale and Unreal Engine assets,
providing photorealistic appearance and rich geometry for learning cross-lens
multi-camera representations.

\begin{figure}[t]
    \centering
     \includegraphics[width=\linewidth]{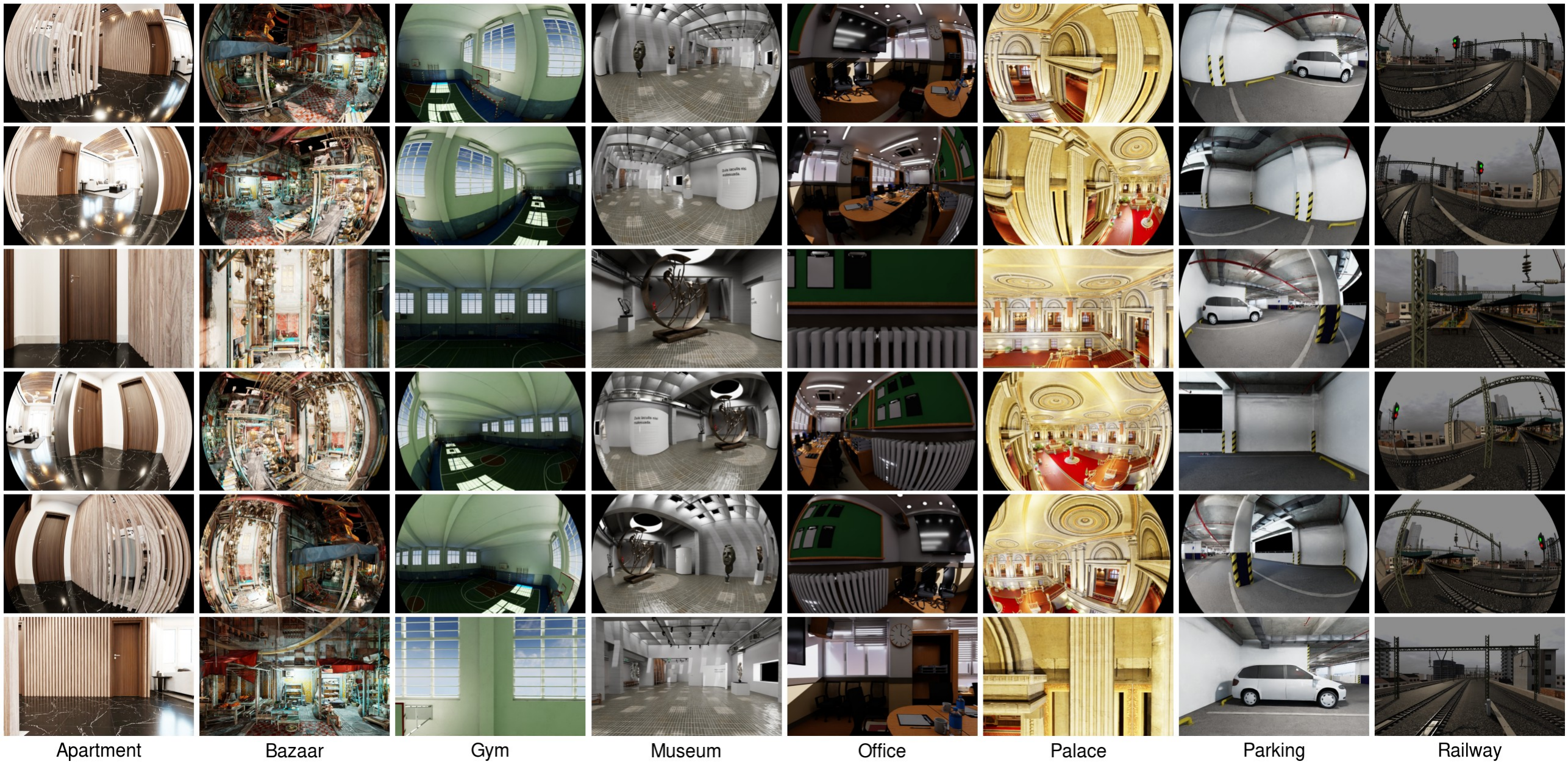}
    \caption{Overview of OmniScene. The dataset covers diverse scene categories, including urban, nature, public, and sci-fi environments.}
    \label{fig:dataset_overview}
\end{figure}

To support these goals at scale, OmniScene contains \textbf{103} distinct
complex scenes, \textbf{564} randomized motion sequences, and approximately
\textbf{266K} multi-view frames, corresponding to over \textbf{1.7M} individual
images across the six-camera rig. Each frame is paired with dense, noise-free
metric ground truth, with an overview shown in \Cref{fig:dataset_overview}.

\paragraph{Heterogeneous Camera Setup.}
Each sequence uses a rigid six-camera sensor suite designed for
near-omnidirectional coverage. Four fisheye cameras are arranged around the platform center with overlapping fields of view to capture horizontal surround context. They follow the
Kannala-Brandt projection model and provide a $180^\circ$ field of view. Two
front- and rear-facing pinhole cameras with standard perspective intrinsics
complement the fisheye views by capturing long-range details. All six cameras are rendered
synchronously at $504 \times 798$ resolution and share a unified rig coordinate
frame. We provide intrinsics and extrinsics for every camera and every frame under
the standard OpenCV convention, allowing models to use the calibrated geometry
directly without unwarping or manual rectification.

\paragraph{Trajectory Generation.}
To obtain diverse ego-motion and viewpoint coverage, OmniScene generates
trajectories with an occupancy-aware constrained waypoint sampler rather than
predefined linear or axis-aligned paths. For each scene, we first derive a
navigable waypoint set from the occupancy representation, retaining only camera
locations with sufficient obstacle clearance and valid geometry. Trajectories are
sampled as constrained random walks over this waypoint set, with bounds on
horizontal displacement, vertical variation, and heading change to ensure smooth
motion while preserving three-dimensional exploration. To avoid locally trapped or
degenerate paths, the sampler adaptively expands the search range, applies
large-angle reorientation, and relocates the rig to more open regions when
progress becomes insufficient. Each trajectory is then validated using a
sliding-window spatial-extent criterion; paths with limited coverage are rejected
and resampled. For each scene, we generate up to 10 independent trajectories of
500 to 1000 frames. The virtual multi-camera rig is actuated along these paths,
with all cameras rendered at synchronized timestamps, so RGB images, depth maps,
semantic annotations, and camera poses are temporally and geometrically aligned by
construction.

\paragraph{Annotations and Quality Control.}
OmniScene provides dense per-pixel supervision for every viewpoint, including
pixel-accurate orthogonal $z$-depth, validity masks, and sky indicators for
invalid or unbounded regions. The post-processing pipeline removes degenerate
frames caused by camera-geometry collisions or uninformative views, masks fisheye
pixels outside the $180^\circ$ optical boundary, isolates sky regions with
unbounded depth, and verifies cross-view depth consistency before export. The
training, validation, and test sets are split by disjoint scenes, enabling
evaluation on unseen environments.
\section{Experiments}
\label{sec:experiments}

\begin{figure}[!t]
\centering
\includegraphics[width=\textwidth]{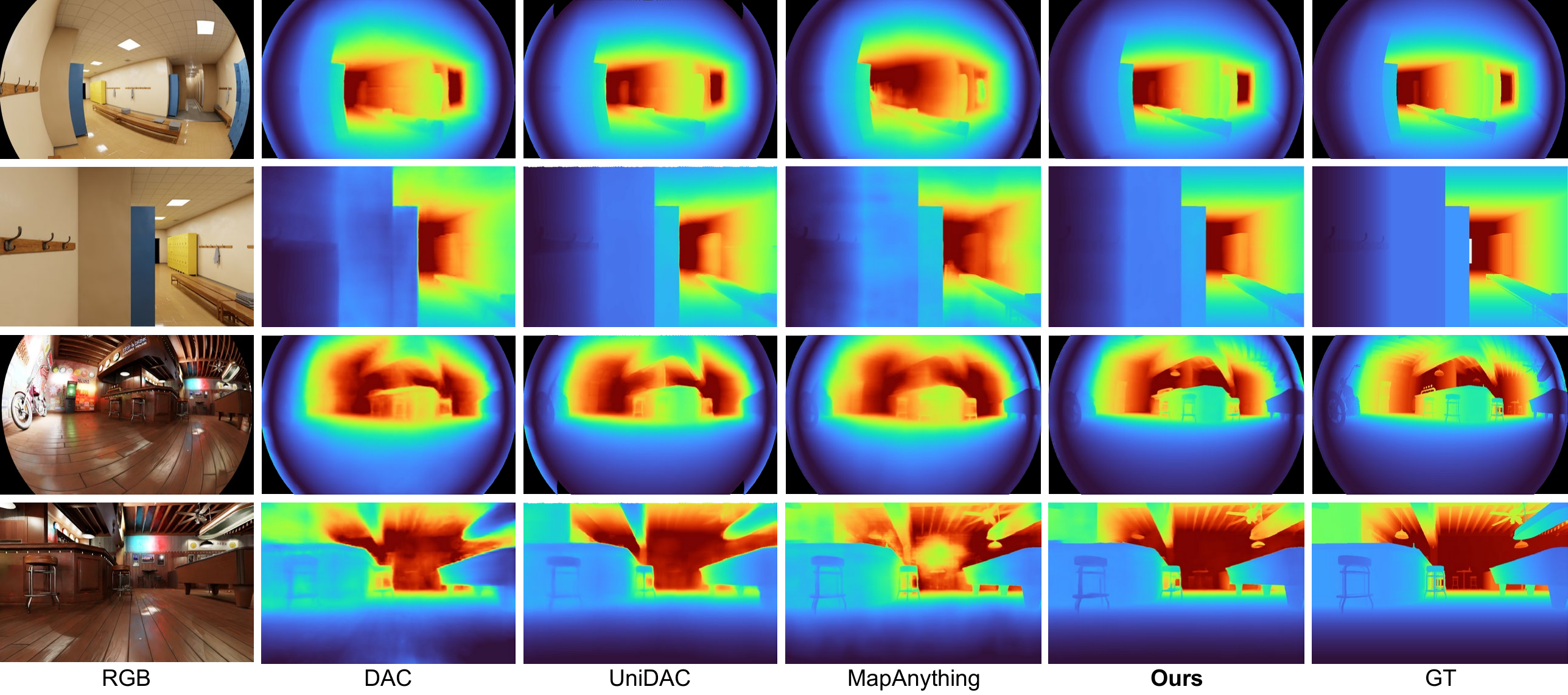}
\caption{Qualitative results from one fisheye view and one pinhole view in the six-view \omniscenefull{} setting.}
\label{fig:syn_results}
\end{figure}

\paragraph{Implementation Details.}
\begin{sloppypar}
We train X-Lens on 14 datasets covering indoor, outdoor, synthetic, and in-the-wild scenes. Stage 1 uses pinhole data from BlendedMVS \cite{yao2020blendedmvs}, Mapillary Planet-Scale Depth \cite{lopezantequera2020mapillary}, ScanNet++ v2 \cite{yeshwanth2023scannetpp}, Spring \cite{mehl2023spring}, TartanAirV2-WB \cite{patel2025tartanground}, UnrealStereo4K \cite{tosi2021smd}, Aria Synthetic Environments \cite{avetisyan2024scenescript}, DL3DV \cite{ling2024dl3dv}, Dynamic Replica \cite{karaev2023dynamicstereo}, MegaDepth \cite{li2018megadepth}, MVS-Synth \cite{huang2018deepmvs}, ParallelDomain-4D \cite{vanhoorick2024gcd}, and SAIL-VOS 3D \cite{hu2021sailvos3d}. Stage 2 uses only the fisheye subset of OmniScene, and Stage 3 jointly uses the Stage-1 pinhole data, the six-camera heterogeneous OmniScene data, and KITTI360~\cite{liao2022kitti360}.
\end{sloppypar}

In Stage 1, we train the full model on pinhole data for 150K steps using 64 H100 GPUs, a 5K-step warm-up, and a peak learning rate of $1\times10^{-4}$. The training resolution is randomly sampled from $504\times504$, $504\times378$, $504\times336$, $504\times280$, $336\times504$, $896\times504$, $756\times504$, and $672\times504$, with the number of views uniformly sampled from $[2,16]$ for square inputs. In Stage 2, we freeze the backbone and DPT decoder and train only the calibration tokens on fisheye data for 10K steps with 500 warm-up steps and a peak learning rate of $5.0\times10^{-5}$. We inject 16 calibration tokens per layer, use a fixed $504\times798$ resolution, and keep the input view count fixed to 4. In Stage 3, we jointly fine-tune on heterogeneous multi-view data and pinhole data for 50K steps using 64 H100 GPUs, a 1K-step warm-up, a peak MLP learning rate of $3.0\times10^{-5}$, a peak Jacobian-bias learning rate of $1.0\times10^{-4}$, and a peak backbone and DPT learning rate of $3.0\times10^{-6}$. The resolution is fixed to $504\times798$, the number of views is randomly sampled from $[6,8]$, and pose conditioning is randomly activated with probability 0.7 during training. We randomly sample camera views during training instead of always using the canonical dataset layouts, improving robustness to unseen multi-camera configurations.

\begin{table}[!tp]
\centering
\caption{Fisheye evaluation on monocular KITTI360~\cite{liao2022kitti360}, monocular \omniscenesingle, and four-view \omniscenequad. FPS is reported at the native evaluation resolution of each dataset. The best result on each dataset is highlighted in bold.}
\label{tab:fisheye_eval}
\footnotesize
\setlength{\tabcolsep}{1.8pt}
\begin{tabular*}{\textwidth}{@{\extracolsep{\fill}}lclccccccc@{}}
\toprule
Dataset & Views & Method & Params & Scale AbsRel $\downarrow$ & AbsRel $\downarrow$ & RMSE $\downarrow$ & $\delta_1$ $\uparrow$ & $\tau_{1.03}$ $\uparrow$ & FPS $\uparrow$ \\
\midrule
\multirow{7}{*}{KITTI360~\cite{liao2022kitti360}} & \multirow{7}{*}{1} & UniDepthv2-Small~\cite{piccinelli2025unidepthv2} & 0.03B & 0.1597 & 0.2597 & 7.4067 & 0.6931 & 0.1803 & 7 \\
 &  & UniDepthv2-Large~\cite{piccinelli2025unidepthv2} & 0.35B & 0.1787 & 0.2718 & 8.0432 & \textbf{0.7091} & \textbf{0.1860} & 5 \\
 &  & Metric3Dv2-Small~\cite{hu2024metric3dv2} & 0.03B & 0.2631 & 0.2690 & 6.4410 & 0.6132 & 0.1477 & 8 \\
 &  & Metric3Dv2-Giant~\cite{hu2024metric3dv2} & 1.38B & 0.1598 & 0.2258 & 6.2469 & 0.6865 & 0.1434 & 4 \\
 &  & DepthAnyCamera~\cite{zhang2025depthanycamera} & 0.06B & 0.1453 & 0.2165 & 4.3103 & 0.6742 & 0.1307 & 8 \\
 &  & UniDAC~\cite{ganesan2026unidac} & 0.36B & 0.1176 & 0.2413 & 4.4671 & 0.6967 & 0.1251 & 6 \\
 &  & Ours & 0.04B & \textbf{0.0955} & \textbf{0.2110} & \textbf{4.1961} & 0.6959 & 0.1436 & \textbf{41} \\
\midrule
\multirow{7}{*}{KITTI360~\cite{liao2022kitti360}} & \multirow{7}{*}{2} & UniDepthv2-Small~\cite{piccinelli2025unidepthv2} & 0.03B & 0.1592 & 0.2595 & 7.4066 & 0.6942 & 0.1807 & 3 \\
 &  & UniDepthv2-Large~\cite{piccinelli2025unidepthv2} & 0.35B & 0.1781 & 0.2699 & 8.0433 & 0.7097 & 0.1868 & 2 \\
 &  & Metric3Dv2-Small~\cite{hu2024metric3dv2} & 0.03B & 0.2633 & 0.2691 & 6.4408 & 0.6132 & 0.1435 & 4 \\
 &  & Metric3Dv2-Giant~\cite{hu2024metric3dv2} & 1.38B & 0.1598 & 0.2257 & 6.2469 & 0.6866 & 0.1434 & 2 \\
 &  & DepthAnyCamera~\cite{zhang2025depthanycamera} & 0.06B & 0.1454 & 0.2163 & 4.3103 & 0.6743 & 0.1307 & 4 \\
 &  & UniDAC~\cite{ganesan2026unidac} & 0.36B & 0.1175 & 0.2413 & 4.4673 & 0.6968 & 0.1254 & 3 \\
 &  & Ours & 0.04B & \textbf{0.0832} & \textbf{0.2069} & \textbf{4.0433} & \textbf{0.7014} & \textbf{0.1557} & \textbf{41} \\
\midrule
\multirow{7}{*}{\omniscenesingle} & \multirow{7}{*}{1} & UniDepthv2-Small~\cite{piccinelli2025unidepthv2} & 0.03B & 0.3793 & 0.1341 & 4.3389 & 0.8771 & 0.2876 & 6 \\
 &  & UniDepthv2-Large~\cite{piccinelli2025unidepthv2} & 0.35B & 0.3263 & 0.1583 & 12.6289 & \textbf{0.9102} & \textbf{0.4775} & 5 \\
 &  & Metric3Dv2-Small~\cite{hu2024metric3dv2} & 0.03B & 0.3607 & 0.2409 & 2.6357 & 0.6931 & 0.1698 & 6 \\
 &  & Metric3Dv2-Giant~\cite{hu2024metric3dv2} & 1.38B & 0.2811 & 0.1352 & 2.2612 & 0.8453 & 0.3541 & 3 \\
 &  & DepthAnyCamera~\cite{zhang2025depthanycamera} & 0.06B & 0.2681 & 0.1963 & 2.8911 & 0.7950 & 0.2771 & 8 \\
 &  & UniDAC~\cite{ganesan2026unidac} & 0.36B & 0.2719 & 0.1869 & 2.8903 & 0.7693 & 0.2691 & 5 \\
 &  & Ours & 0.04B & \textbf{0.2287} & \textbf{0.1191} & \textbf{1.9801} & 0.8868 & 0.3357 & \textbf{34} \\
\midrule
\multirow{4}{*}{\omniscenequad} & \multirow{4}{*}{4} & MapAnything~\cite{keetha2025mapanything} & 1.23B & 1.0455 & 0.1918 & 2.1602 & 0.6827 & 0.1178 & 6 \\
 &  & DepthAnyCamera~\cite{zhang2025depthanycamera} & 0.06B & 0.2681 & 0.1963 & 2.8911 & 0.7950 & 0.2771 & 2 \\
 &  & UniDAC~\cite{ganesan2026unidac} & 0.36B & 0.2719 & 0.1869 & 2.8903 & 0.7693 & 0.2691 & 1 \\
 &  & Ours & 0.04B & \textbf{0.1268} & \textbf{0.1138} & \textbf{1.6481} & \textbf{0.8822} & \textbf{0.3328} & \textbf{24} \\
\bottomrule
\end{tabular*}
\vspace{0.2em}
\caption*{KITTI360~\cite{liao2022kitti360} is evaluated at $504\times504$ resolution, while \omniscenequad{} and \omniscenesingle{} are evaluated at $504\times798$ resolution.}
\end{table}

\paragraph{Baseline Methods.}
\begin{sloppypar}
We compare against representative monocular, wide-FOV, and feed-forward multi-view baselines according to the camera setting of each benchmark. For fisheye and heterogeneous-camera evaluation, we include UniDepthv2 \cite{piccinelli2025unidepthv2}, Metric3Dv2 \cite{hu2024metric3dv2}, DepthAnyCamera \cite{zhang2025depthanycamera}, UniDAC \cite{ganesan2026unidac}, and MapAnything \cite{keetha2025mapanything}. For pinhole multi-view evaluation, we further compare with VGGT \cite{wang2025vggt}, VGGT-Omega \cite{wang2026vggtomega}, DepthAnything3 \cite{lin2025depthanything3}, and MapAnything \cite{keetha2025mapanything}. DepthAnyCamera \cite{zhang2025depthanycamera} and UniDAC \cite{ganesan2026unidac} are monocular any-camera depth estimators, so in multi-view settings we apply them independently to each input image and do not perform cross-view fusion. For MapAnything \cite{keetha2025mapanything} and DepthAnything3 \cite{lin2025depthanything3}, we directly provide the available camera intrinsics and extrinsics in our experiments. For methods that do not explicitly predict a global scale term, we mark the Scale AbsRel entry with a red cross.
\end{sloppypar}

\paragraph{Datasets \& Metric.}
We evaluate homogeneous fisheye, homogeneous pinhole, and heterogeneous-camera settings. For fisheye evaluation, KITTI360~\cite{liao2022kitti360} is used in both monocular and two-view settings, where the two-view input is formed by selecting two frames separated by 6 timestamps along the same trajectory. We also evaluate monocular \omniscenesingle{} and four-view \omniscenequad. KITTI360~\cite{liao2022kitti360} is evaluated at $504\times504$ resolution, while \omniscenequad{} and \omniscenesingle{} are evaluated at $504\times798$ resolution. For pinhole evaluation, we use ETH3D~\cite{schops2017eth3d} and ScanNet++V2~\cite{yeshwanth2023scannetpp} with two input views and OmniOcc with six input views~\cref{sec:appendix_omniocc}. For heterogeneous evaluation, \omniscenefull{} contains six input views composed of four fisheye and two pinhole cameras and is evaluated at $504\times798$ resolution. We report Scale AbsRel, AbsRel, RMSE, $\delta_1$, $\tau_{1.03}$, and FPS. Scale AbsRel measures the error of the predicted global scale, while AbsRel, RMSE, and $\delta_1$ follow standard depth-estimation protocols. $\tau_{1.03}$ measures the fraction of pixels whose relative depth error is within 3\%. For every benchmark we adopt scene-disjoint train/validation/test splits so that no evaluation scene is seen during training. The full split protocol for OmniScene and all evaluation datasets are detailed in the supplementary material~\cref{sec:appendix_pinhole_replay}. All FPS numbers are measured on a single NVIDIA H100 GPU.

\subsection{Homogeneous Cameras}

\begin{figure}[!t]
\centering
\includegraphics[width=\textwidth]{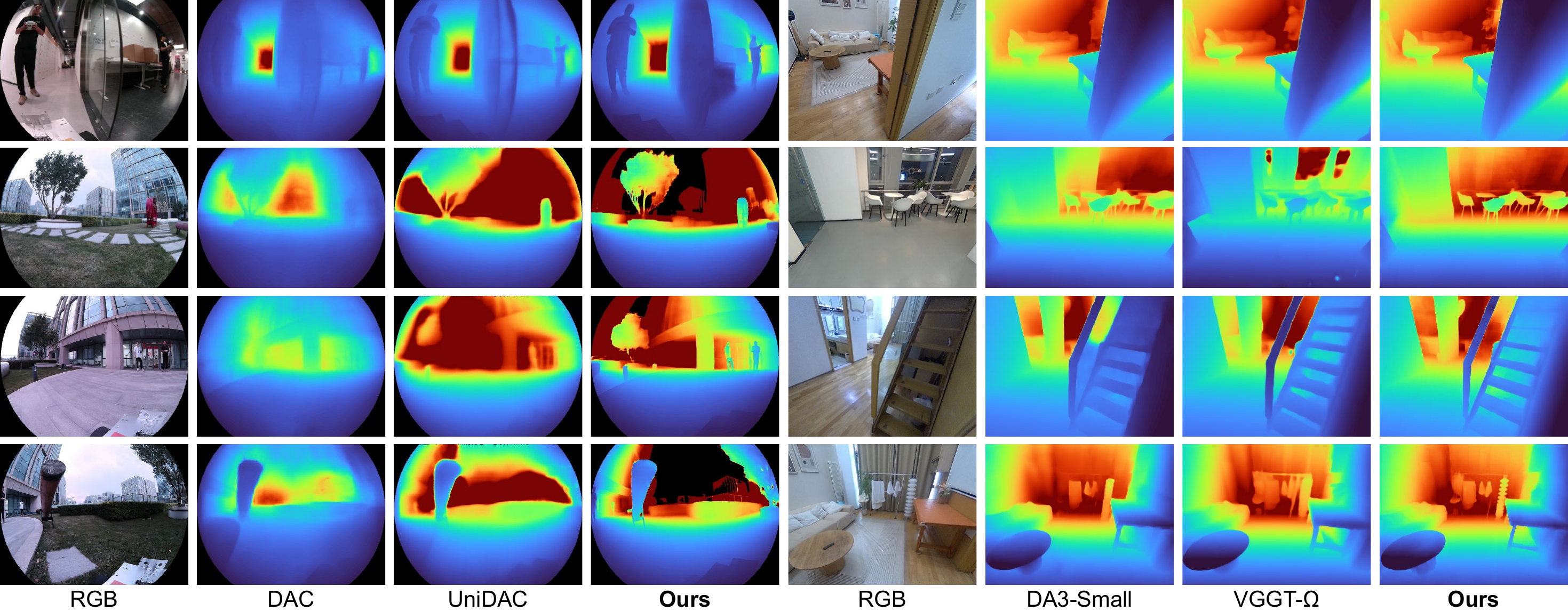}
\caption{Qualitative results on the real-world OmniOcc dataset and real-world fisheye scenes using \ours. The figure shows fisheye-only and pinhole-only results across outdoor and indoor scenes.}
\label{fig:real_results}
\end{figure}

We first evaluate whether \ours preserves strong metric-depth performance under homogeneous camera setups. These experiments cover both fisheye and pinhole inputs and assess whether the model handles wide-FOV distortion efficiently while remaining competitive with large feed-forward geometry models under a much smaller parameter budget. We provide qualitative examples in \Cref{fig:real_results} to visualize depth sharpness and scale consistency on real fisheye-only and pinhole-only inputs.

\paragraph{Fisheye Cameras}

\Cref{tab:fisheye_eval} reports fisheye-camera evaluation across monocular KITTI360~\cite{liao2022kitti360}, two-view KITTI360~\cite{liao2022kitti360}, \omniscenesingle, and \omniscenequad. On monocular KITTI360~\cite{liao2022kitti360}, \ours achieves the best Scale AbsRel, AbsRel, and RMSE among all compared methods, while running at 41 FPS, substantially faster than baseline approaches. When two temporally separated KITTI360~\cite{liao2022kitti360} frames are used as input, \ours further improves over its single-view variant, lowering Scale AbsRel by 12.9\% and RMSE by 3.6\%, indicating that cross-view constraints provide useful geometric evidence beyond monocular inference. On \omniscenesingle, \ours obtains the best Scale AbsRel, AbsRel, and RMSE, and on \omniscenequad{} it consistently outperforms MapAnything~\cite{keetha2025mapanything}, DepthAnyCamera~\cite{zhang2025depthanycamera}, and UniDAC~\cite{ganesan2026unidac} across all reported accuracy metrics while maintaining 24 FPS. Compared with MapAnything~\cite{keetha2025mapanything} on \omniscenequad, \ours uses 96.7\% fewer parameters. These results show that \ours can process raw fisheye views efficiently without relying on panorama stitching or hardware-specific stereo volumes. The fisheye-only examples in \Cref{fig:real_results} show that \ours preserves coherent near-field depth and wide-FOV structure without panorama stitching.

\paragraph{Pinhole Cameras}

\Cref{tab:pinhole_multiview} evaluates pinhole multi-view depth estimation on ETH3D~\cite{schops2017eth3d} and ScanNet++V2~\cite{yeshwanth2023scannetpp} at 504 x 588 resolution and OmniOcc at 504 x 798 resolution. Large geometric foundation models such as VGGT~\cite{wang2025vggt}, VGGT-Omega~\cite{wang2026vggtomega}, DepthAnything3~\cite{lin2025depthanything3}, and MapAnything~\cite{keetha2025mapanything} often achieve strong dense-depth accuracy, but they rely on substantially larger parameter budgets and do not always provide an explicit global scale prediction. In contrast, \ours uses only 0.04B parameters and directly predicts the global scale term, achieving the best Scale AbsRel on ETH3D~\cite{schops2017eth3d} and OmniOcc and competitive performance on ScanNet++V2~\cite{yeshwanth2023scannetpp}. Relative to DA3-Giant~\cite{lin2025depthanything3}, the strongest pinhole baseline on several dense-depth metrics, \ours reduces the parameter count by 97.1\%. Although large geometric foundation models remain strong on some dense metrics, \ours runs much faster than most large feed-forward geometry baselines, reaching 39 FPS on two-view pinhole inputs and 26 FPS on six-view OmniOcc. These results support our design goal of decoupling depth perception from full reconstruction to obtain a compact model suitable for real-time deployment. The pinhole-only examples in \Cref{fig:real_results} further illustrate that \ours maintains metric depth consistency on higher-resolution long-range views.

\begin{table}[t]
\centering
\caption{Multi-view pinhole evaluation on ETH3D~\cite{schops2017eth3d}, ScanNet++V2~\cite{yeshwanth2023scannetpp}, and OmniOcc six-view dataset. The best result on each dataset is highlighted in bold. A red cross indicates that the corresponding baseline does not explicitly predict a global scale term.}
\label{tab:pinhole_multiview}
\footnotesize
\setlength{\tabcolsep}{1.8pt}
\begin{tabular*}{\textwidth}{@{\extracolsep{\fill}}lclccccccc@{}}
\toprule
Dataset & Views & Method & Params & Scale AbsRel $\downarrow$ & AbsRel $\downarrow$ & RMSE $\downarrow$ & $\delta_1$ $\uparrow$ & $\tau_{1.03}$ $\uparrow$ & FPS $\uparrow$ \\
\midrule
\multirow{6}{*}{ETH3D~\cite{schops2017eth3d}} & \multirow{6}{*}{2} & VGGT~\cite{wang2025vggt} & 1.26B & \textcolor{red}{$\times$} & 0.0184 & 0.4073 & 0.9974 & 0.8693 & 15 \\
 &  & VGGT-Omega~\cite{wang2026vggtomega} & 1.14B & \textcolor{red}{$\times$} & \textbf{0.0055} & \textbf{0.1465} & \textbf{0.9994} & 0.8693 & 12 \\
 &  & DA3-Small~\cite{lin2025depthanything3} & 0.03B & \textcolor{red}{$\times$} & 0.0454 & 0.7438 & 0.9501 & 0.5502 & 41 \\
 &  & DA3-Giant~\cite{lin2025depthanything3} & 1.36B & \textcolor{red}{$\times$} & 0.0113 & 0.2081 & 0.9989 & \textbf{0.9073} & 11 \\
 &  & MapAnything~\cite{keetha2025mapanything} & 1.23B & 0.1410 & 0.0228 & 0.3743 & 0.9992 & 0.7800 & 13 \\
 &  & Ours & \textbf{0.04B} & \textbf{0.1217} & 0.0445 & 0.6991 & 0.9723 & 0.5552 & 39 \\
\midrule
\multirow{6}{*}{ScanNet++V2~\cite{yeshwanth2023scannetpp}} & \multirow{6}{*}{2} & VGGT~\cite{wang2025vggt} & 1.26B & \textcolor{red}{$\times$} & 0.0361 & 0.1393 & 0.9753 & 0.7282 & 15 \\
 &  & VGGT-Omega~\cite{wang2026vggtomega} & 1.14B & \textcolor{red}{$\times$} & \textbf{0.0332} & 0.1390 & 0.9756 & \textbf{0.7824} & 12 \\
 &  & DA3-Small~\cite{lin2025depthanything3} & 0.03B & \textcolor{red}{$\times$} & 0.0618 & 0.1754 & 0.9586 & 0.4606 & 41 \\
 &  & DA3-Giant~\cite{lin2025depthanything3} & 1.36B & \textcolor{red}{$\times$} & 0.0354 & \textbf{0.1348} & \textbf{0.9768} & 0.7379 & 11 \\
 &  & MapAnything~\cite{keetha2025mapanything} & 1.23B & \textbf{0.0462} & 0.0548 & 0.1542 & 0.9773 & 0.4523 & 13 \\
 &  & Ours & \textbf{0.04B} & 0.0544 & 0.0549 & 0.1621 & 0.9626 & 0.5423 & 39 \\
\midrule
\multirow{6}{*}{OmniOcc} & \multirow{6}{*}{6} & VGGT~\cite{wang2025vggt} & 1.26B & \textcolor{red}{$\times$} & 0.0458 & \textbf{0.2031} & \textbf{0.9721} & 0.6717 & 9 \\
 &  & VGGT-Omega~\cite{wang2026vggtomega} & 1.14B & \textcolor{red}{$\times$} & 0.0482 & 0.2227 & 0.9703 & 0.6822 & 7 \\
 &  & DA3-Small~\cite{lin2025depthanything3} & 0.03B & \textcolor{red}{$\times$} & 0.0655 & 0.2736 & 0.9311 & 0.4937 & 29 \\
 &  & DA3-Giant~\cite{lin2025depthanything3} & 1.36B & \textcolor{red}{$\times$} & \textbf{0.0437} & 0.2272 & 0.9662 & \textbf{0.7302} & 6 \\
 &  & MapAnything~\cite{keetha2025mapanything} & 1.23B & 0.2050 & 0.0553 & 0.2305 & 0.9628 & 0.5851 & 8 \\
 &  & Ours & \textbf{0.04B} & \textbf{0.0670} & 0.0656 & 0.2584 & 0.9494 & 0.5550 & 26 \\
\bottomrule
\end{tabular*}
\end{table}

\subsection{Heterogeneous Cameras}

\begin{figure}[!tp]
\centering
\includegraphics[width=\textwidth]{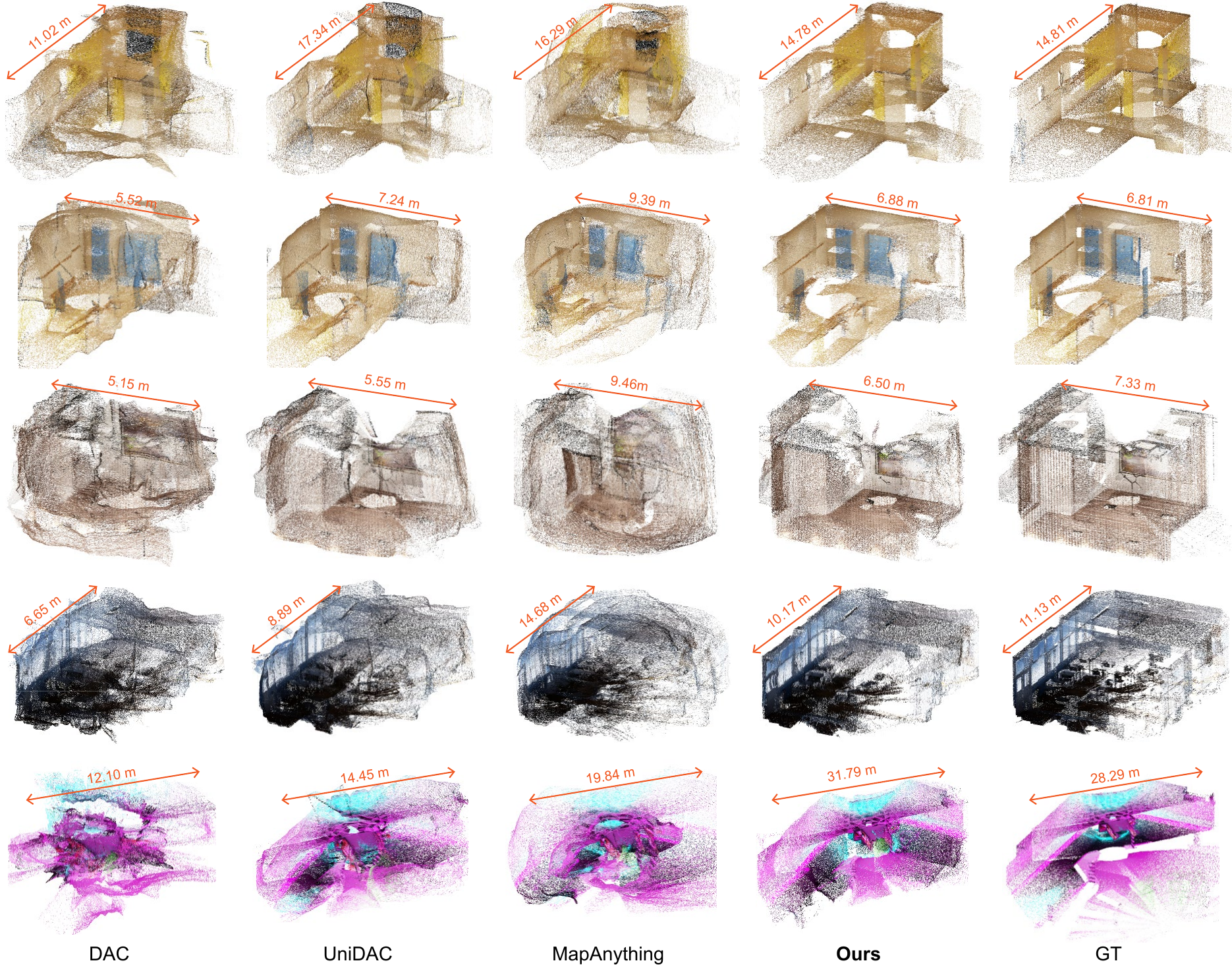}
\caption{Point-cloud comparison on the synthetic \omniscenefull{} setting with six multi-view inputs, consisting of four fisheye and two pinhole views. Compared with DepthAnyCamera~\cite{zhang2025depthanycamera}, UniDAC~\cite{ganesan2026unidac}, and MapAnything~\cite{keetha2025mapanything}, \ours reconstructs more accurate geometry and preserves metric scale consistently.}
\label{fig:pointcloud_syn}
\end{figure}

\begin{table}[!tp]
\centering
\caption{Heterogeneous-camera evaluation \omniscenefull{} with six input views (4 fisheye + 2 pinhole). FPS is measured at $504\times798$ resolution. The best result is highlighted in bold.}
\label{tab:heterogeneous_eval}
\footnotesize
\setlength{\tabcolsep}{1.8pt}
\begin{tabular*}{\textwidth}{@{\extracolsep{\fill}}lclccccccc@{}}
\toprule
Dataset & Views & Method & Params & Scale AbsRel $\downarrow$ & AbsRel $\downarrow$ & RMSE $\downarrow$ & $\delta_1$ $\uparrow$ & $\tau_{1.03}$ $\uparrow$ & FPS $\uparrow$ \\
\midrule
\multirow{4}{*}{\omniscenefull} & \multirow{4}{*}{6} & MapAnything~\cite{keetha2025mapanything} & 1.23B & 0.3701 & 0.1746 & 2.1834 & 0.7357 & 0.1647 & 5 \\
 &  & DepthAnyCamera~\cite{zhang2025depthanycamera} & 0.06B & 0.2571 & 0.2066 & 2.3981 & 0.7653 & 0.2411 & 1 \\
 &  & UniDAC~\cite{ganesan2026unidac} & 0.36B & 0.2506 & 0.1368 & 2.6125 & 0.8156 & 0.2511 & 1 \\
 &  & Ours & 0.04B & \textbf{0.1181} & \textbf{0.1021} & \textbf{1.5993} & \textbf{0.8982} & \textbf{0.3724} & \textbf{22} \\
\bottomrule
\end{tabular*}
\end{table}

We further evaluate \ours under the heterogeneous setting, where fisheye and pinhole cameras are jointly used as input. This setting assesses whether a compact feed-forward model can fuse mixed camera views while preserving metric consistency. As shown in \Cref{tab:heterogeneous_eval}, \ours achieves the best performance on \omniscenefull{} across all reported metrics, with the lowest AbsRel and RMSE. Compared with MapAnything~\cite{keetha2025mapanything}, \ours lowers Scale AbsRel by 68.1\%, uses 96.7\% fewer parameters, and runs more than four times faster. DepthAnyCamera~\cite{zhang2025depthanycamera} and UniDAC~\cite{ganesan2026unidac} are monocular any-camera methods and are therefore applied independently to each view. Without cross-view fusion, they lag behind \ours in both accuracy and speed. These results demonstrate that explicitly modeling heterogeneous multi-view depth provides a practical path toward real-time metric perception on mixed fisheye and pinhole rigs. \Cref{fig:syn_results} visualizes one fisheye view and one pinhole view from the same six-view \omniscenefull{} sample, showing cross-camera depth consistency across near-field wide-FOV observations and long-range pinhole observations. \Cref{fig:pointcloud_syn} further compares multi-view point clouds on \omniscenefull{}, where \ours preserves the scene geometry and metric scale more faithfully than DepthAnyCamera~\cite{zhang2025depthanycamera}, UniDAC~\cite{ganesan2026unidac}, and MapAnything~\cite{keetha2025mapanything}.

\section{Ablation Study}
\label{sec:ablation}

\begin{figure}[!tp]
\centering
\includegraphics[width=\textwidth]{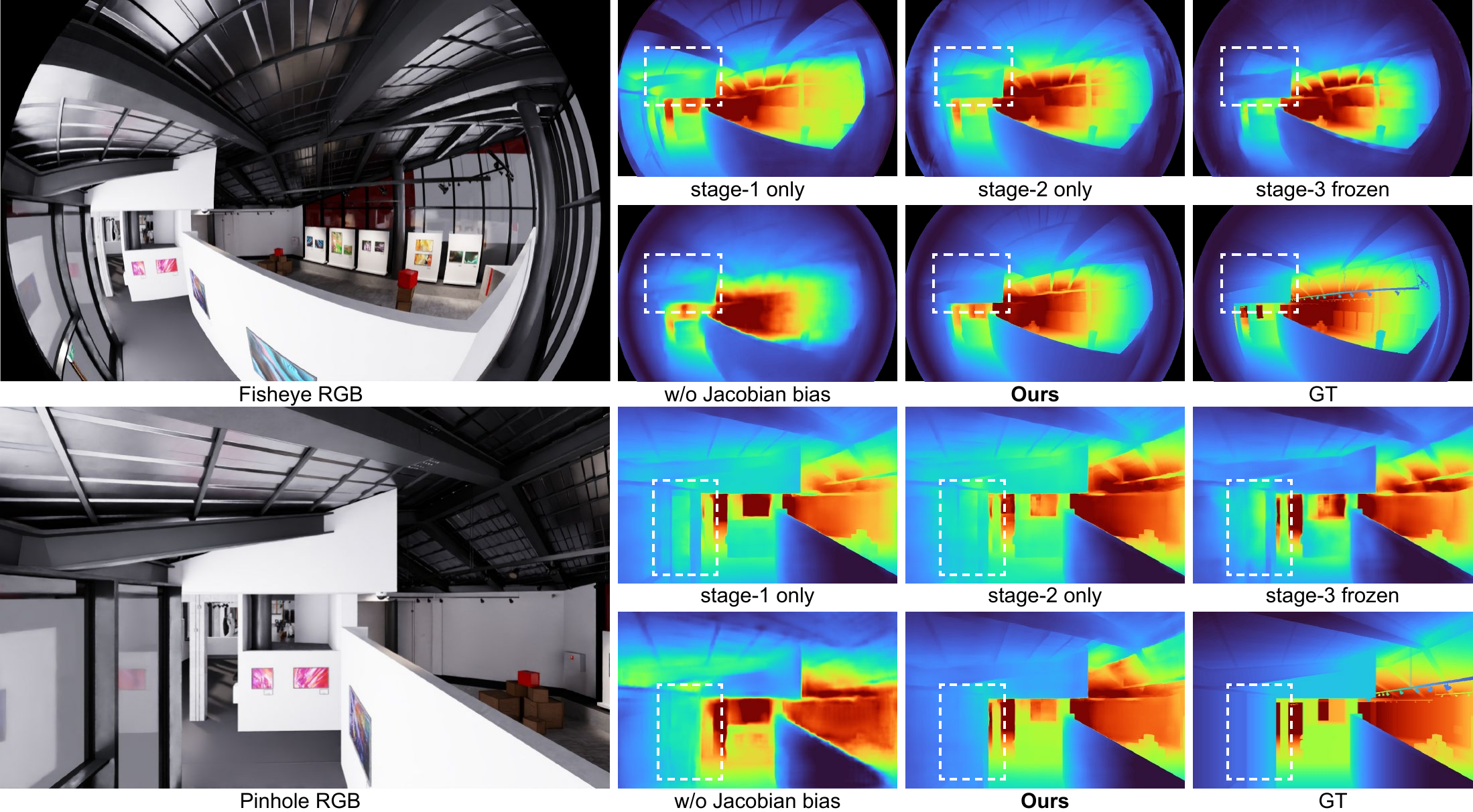}
\caption{Qualitative ablation on the six-view heterogeneous \omniscenefull{} setting. We show predictions from one fisheye view and one pinhole view using the same ablation variants as in \Cref{tab:ablation_training_geometry}.}
\label{fig:ablation_hete}
\end{figure}

\begin{figure}[!tp]
\centering
\includegraphics[width=\textwidth]{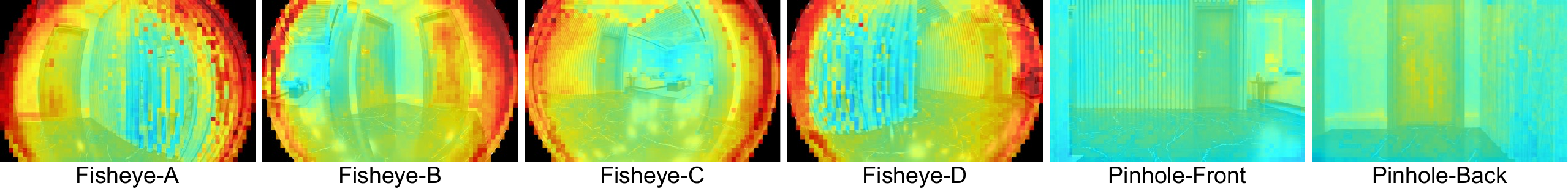}
\caption{Visualization of the bias-correction magnitude from $\mathbf{B}$ in \Cref{eq:attention_bias_formula_en}. Fisheye views require larger corrections in highly distorted boundary regions, while pinhole views show smoother and more spatially uniform corrections, reflecting their approximately linear projection, which does not require rapid spatial changes in the correction field.}
\label{fig:ablation_attention}
\end{figure}

We ablate the main training stages and geometry modules on the heterogeneous \omniscenefull{} setting, the fisheye-only setting, and the pinhole-only OmniOcc setting. Our loss terms follow DepthAnything3~\cite{lin2025depthanything3}, with the local depth consistency term adopted from MoGe~\cite{wang2024moge}. The names of the variants follow the method components. \emph{Stage-1 only} uses only the initial pinhole training stage. \emph{Stage-2 only} adds fisheye calibration-token adaptation but skips heterogeneous joint training. \emph{Stage-3 frozen} freezes the backbone and DPT decoder during Stage 3. \emph{w/o Jacobian bias} removes the Jacobian Distortion Bias and directly fine-tunes the backbone on mixed-camera data. \emph{Ours} uses the full three-stage training pipeline with calibration tokens and Jacobian Distortion Bias.

\begin{table}[t]
\centering
\caption{Ablation of training stages and geometry modules. The heterogeneous setting uses \omniscenefull{} with four fisheye and two pinhole cameras. The fisheye-only setting uses fisheye cameras. The pinhole-only setting uses six-view OmniOcc.}
\label{tab:ablation_training_geometry}
\footnotesize
\setlength{\tabcolsep}{2.0pt}
\begin{tabular*}{\textwidth}{@{\extracolsep{\fill}}lclccccc@{}}
\toprule
Cameras & Views & Variant & Scale AbsRel $\downarrow$ & AbsRel $\downarrow$ & RMSE $\downarrow$ & $\delta_1$ $\uparrow$ & $\tau_{1.03}$ $\uparrow$ \\
\midrule
\multirow{5}{*}{Heterogeneous} & \multirow{5}{*}{6} & Stage-1 only & 0.3026 & 0.7563 & 2.4716 & 0.4811 & 0.1156 \\
 & & Stage-2 only & 0.3308 & 0.3952 & 2.2159 & 0.6264 & 0.1331 \\
 & & Stage-3 frozen & 0.2886 & 0.2333 & 2.6006 & 0.7436 & 0.1743 \\
 & & w/o Jacobian bias & 0.1801 & 0.1912 & 2.5437 & 0.7812 & 0.2102 \\
 & & Ours & \textbf{0.1181} & \textbf{0.1021} & \textbf{1.5993} & \textbf{0.8982} & \textbf{0.3724} \\
\midrule
\multirow{4}{*}{Fisheye} & \multirow{4}{*}{4} & Stage-1 only & 0.7325 & 0.2769 & 2.9182 & 0.6719 & 0.1711 \\
 & & Stage-2 only & 0.2126 & 0.2083 & 2.5082 & 0.7560 & 0.1990 \\
 & & Stage-3 frozen & 0.4018 & 0.2946 & 2.8792 & 0.6805 & 0.1436 \\
 & & w/o Jacobian bias & 0.1854 & 0.2115 & 2.8174 & 0.7717 & 0.1998 \\
 & & Ours & \textbf{0.1268} & \textbf{0.1138} & \textbf{1.6481} & \textbf{0.8822} & \textbf{0.3328} \\
\midrule
\multirow{5}{*}{Pinhole} & \multirow{5}{*}{6} & Stage-1 only & 0.1052 & 0.0717 & 0.2801 & 0.9439 & 0.5433 \\
 & & Stage-2 only & 0.1052 & 0.0717 & 0.2801 & 0.9439 & 0.5433 \\
 & & Stage-3 frozen & 0.1916 & 0.0718 & 0.2774 & 0.9442 & 0.5025 \\
 & & w/o Jacobian bias & 0.5631 & 0.2844 & 0.8767 & 0.5332 & 0.0859 \\
 & & Ours & \textbf{0.0670} & \textbf{0.0656} & \textbf{0.2584} & \textbf{0.9494} & \textbf{0.5551} \\
\bottomrule
\end{tabular*}
\end{table}

\Cref{tab:ablation_training_geometry} shows that each stage contributes to robust metric depth. Stage-1 only performs poorly under mixed cameras, as pinhole-only training cannot handle fisheye distortion. Stage-2 only improves dense depth but remains limited without heterogeneous joint optimization. Stage-3 frozen also underperforms, indicating that the shared representation must adapt to mixed camera geometry. Removing the Jacobian Distortion Bias degrades results across the evaluated camera settings, showing that the bias improves cross-view geometric alignment and also regularizes generalization when training mixes heterogeneous and pinhole data. The full model achieves the best results across settings with complete results. \Cref{fig:ablation_hete} visualizes the qualitative effect of the same ablation variants on one fisheye view and one pinhole view from the six-view heterogeneous setting. \Cref{fig:ablation_attention} further visualizes the magnitude of the bias correction $\mathbf{B}$ in \Cref{eq:attention_bias_formula_en}. Fisheye views require stronger corrections near high-distortion boundary regions, whereas pinhole views show smoother and more spatially uniform corrections, reflecting their approximately linear projection, which does not require rapid spatial changes in the correction field.
\section{Conclusion}
\label{sec:conclusion}
We present \ours, a compact feed-forward model for real-time metric depth estimation with heterogeneous fisheye and pinhole cameras. By focusing the prediction target on dense depth and a global metric scale, and by introducing calibration tokens together with a Jacobian-parameterized distortion bias, \ours reconciles camera-dependent projection differences while preserving native image geometry. Extensive experiments show that this lightweight 0.04B-parameter design achieves superior heterogeneous-camera depth accuracy, reducing AbsRel by 25.4\% over the strongest baseline while using 88.9\% fewer parameters, and remains effective on fisheye-only and pinhole-only settings. We also introduce OmniScene, a large-scale synthetic dataset with approximately 266K synchronized six-view frames across diverse indoor and outdoor scenarios, and demonstrate the potential of \ours as a geometry-aware visual encoder for downstream Vision-Action models.

\paragraph{Limitations.} X-Lens dynamically accepts arbitrary intrinsic and extrinsic parameters as geometric inputs; however, it strictly relies on ground-truth calibration and lacks the capability to jointly predict these configurations online. Regarding generalization, while the system generalizes across diverse camera setups, the underlying synthetic training data inherently covers a bounded range of camera intrinsic variations. Consequently, when encountering unseen fisheye lens models with extreme FOV configurations that deviate drastically from the training distribution, a noticeable sim-to-real gap remains, causing a slight degradation in performance. Future work will broaden real-data and intrinsic diversity, add temporal aggregation, and explore X-Lens as a geometric prior for world models, SLAM, and robotic loco-manipulation.

{
    \small
    \bibliographystyle{ieeenat_fullname}
    \bibliography{main}
}

\clearpage
\appendix
\renewcommand{\thepage}{S\arabic{page}}
\setcounter{page}{1}

\section{Applications in Vision-Action Models}
\label{sec:vla}
In embodied AI and robotic manipulation, Vision-Action models have emerged as a dominant paradigm for generating precise, end-to-end control policies. However, state-of-the-art VA frameworks typically rely on 2D visual foundation models, such as CLIP~\cite{radford2021clip} or DINOv3~\cite{Simeoni2025DINOv3}, as their primary encoders. While these backbones excel at extracting high-level semantic abstractions, 2D foundation models are not explicitly optimized for metric geometry; spatial attributes such as absolute metric scale, 3D depth continuity, and surface geometry are therefore not directly supervised during standard 2D self-supervised pretraining. Consequently, the downstream policy head may need to recover task-relevant 3D spatial relations from limited robotic trajectory data.

To bridge this gap, we extend our proposed \ours as a geometry-infused visual encoder for VA models. Unlike pure 2D semantic representations, our approach provides two practical properties for downstream policy learning: (a) Explicit Geometric Grounding: Our backbone is explicitly optimized for multi-view geometry and pixel-aligned depth estimation, encouraging the latent features to encode 3D structure of the physical environment. (b) Information-Rich Routing: Rather than feeding rigid, reconstructed 3D point clouds or explicit depth maps that may suffer from quantization errors and missing semantic details, we directly route flexible self-attention tokens, preserving both task semantics and spatial priors for downstream task success rate evaluation.

\subsection{Integration Pipeline}
To enable a controlled architectural comparison between fundamentally disparate visual foundation models, we establish a unified modular encoder interface contract that strictly decouples the perception backbone $\mathcal{E}$ from the downstream fusion head $g$. Formally, at each control timestep, the robot collects a multi-view observation set $\mathbf{I} = \{I_v\}_{v=1}^{V}$ ($I_v \in \mathbb{R}^{3 \times H \times W}$) and a 14-dimensional proprioceptive state $\mathbf{q}$, which are processed via a composite vision frontend to yield a conditioning embedding $\mathbf{z} = (g \circ \mathcal{E})(\mathbf{I})$ for the conditional flow-matching~\cite{lipman2023flowmatching} DiT~\cite{peebles2023dit} policy $\pi_\theta(\mathbf{a} \mid \mathbf{z}, \mathbf{q})$. Under our interface contract, any compliant backbone must return a canonical token dictionary containing 2D image patch tokens $T_{\text{img}}$, cached positions, and per-layer camera ($\mathcal{C}_\ell$) or spatial ($\mathcal{S}_\ell$) tokens. In the baseline branch, we instantiate a frozen DINOv3-ViT-B backbone which populates only $T_{\text{img}}$ while setting $\mathcal{C}_\ell = \mathcal{S}_\ell = \emptyset$, collapsing $g$ to a vanilla 2D semantic representation. Conversely, in our proposed branch, we introduce a drop-in backbone substitution using our pre-trained X-Lens model. Architecturally, \ours utilizes a ViT backbone featuring nested, alternating frame-attention blocks and global-attention blocks. The intermediate aggregator features are decoupled via dual-stream splitting into an interleaved frame appearance stream and a cross-view global geometry stream, with the latter populating $\mathcal{S}_\ell$. This spatial token is split into local and global segments, projected via separate MLPs, and concatenated into a unified embedding $S' = \left[ \text{MLP}_{\text{local}}(S) \;\parallel\; \text{MLP}_{\text{global}}(S) \right] \in \mathbb{R}^{1024}$, which is compressed by a ResNetAdapter into a $2 \times 2 \times 1024$ map $\mathbf{z}_{\text{vis}}$ and flattened with $\mathbf{q}$ to synthesize $\mathbf{z}$. 

\subsection{Experiments}
\label{sec:exp_on_vla}
To further assess whether the geometric representation can benefit policy learning, we conduct a preliminary case study on the RoboTwin2~\cite{Chen2025RoboTwin2A} Scan-Object (Easy) task. For this single-task comparison, we evaluate our method against several representative visual encoders, including DINOv3-Base, DINOv3-Large~\cite{Simeoni2025DINOv3}, and VGGT~\cite{wang2025vggt}.

Furthermore, we benchmark our framework against VO-DPP, an advanced, proprietary extension of the original VO-DP~\cite{Ni2025VODPSA} algorithm. While VO-DP is inherently limited to single-view configurations, VO-DPP introduces native support for multi-view feed-forward inputs, establishing a strong baseline for multi-camera geometric perception. Although VO-DPP remains unreleased to the public at the time of this writing, evaluating against this enhanced variant provides an initial indication of our model's downstream task success rate in a multi-view robotic scanning scenario.

\begin{table}[h]
\centering
\caption{Preliminary case study of different visual representation algorithms combined with VO-DPP on RoboTwin2 Scan-Object (Easy) downstream task success rate.}
\label{tab:appendix_va}
\footnotesize
\setlength{\tabcolsep}{2.0pt}
\begin{tabular*}{\textwidth}{@{\extracolsep{\fill}}lcccc@{}}
\toprule
Variant & visual encoder & epoch & learning\_rate & TSR(\%)  \\
\midrule
With dinov3-base & training & 300 & 1e-5 & 50  \\
With dinov3-large & training & 300 & 1e-5 & 55  \\
With VGGT & training & 300 & 1e-5 & 67  \\
With Ours & training & 300 & 1e-5 & 72  \\
\bottomrule
\end{tabular*}
\end{table}
\section{Additional Ablation Studies}
\label{sec:appendix_ablation}

\subsection{Auxiliary Camera-Head Prediction}
\label{sec:appendix_camera_head}

We further test whether adding explicit camera-parameter prediction helps the model. This auxiliary variant follows the first-stage pinhole training setup and adds a camera head with learnable tokens to predict intrinsics and extrinsics. It also predicts a ray map with an additional DPT head and optimizes the ray map together with depth through a point loss. The training schedule, data, and parameter budget are otherwise kept aligned with the main Stage-1 setup.

\begin{table}[h]
\centering
\caption{Auxiliary camera-head experiment on six-view OmniOcc after Stage-1 training. Adding explicit camera and ray-map prediction hurts metric depth accuracy.}
\label{tab:appendix_camera_head}
\footnotesize
\setlength{\tabcolsep}{2.0pt}
\begin{tabular*}{\textwidth}{@{\extracolsep{\fill}}lccccc@{}}
\toprule
Variant & Scale AbsRel $\downarrow$ & AbsRel $\downarrow$ & RMSE $\downarrow$ & $\delta_1$ $\uparrow$ & $\tau_{1.03}$ $\uparrow$ \\
\midrule
With camera head & 0.1816 & 0.0702 & 0.2861 & 0.8961 & 0.4910 \\
Ours after Stage-1 training & \textbf{0.0692} & \textbf{0.0636} & \textbf{0.2499} & \textbf{0.9481} & \textbf{0.5442} \\
\bottomrule
\end{tabular*}
\end{table}

The auxiliary camera head increases coupling between depth, ray maps, and camera parameters. This hurts depth accuracy despite using the same training data and comparable training steps. The result supports our design choice of keeping camera information on the input side and restricting the externally visible outputs to depth, confidence, and global scale.

\subsection{Effect of Pure-Pinhole Data in Stage 3}
\label{sec:appendix_pinhole_replay}

In \cref{sec:method:training_pipeline}, we mix pure-pinhole multi-view data into Stage 3 because OmniScene's two pinhole views barely overlap, which would otherwise erode the model's multi-view pinhole performance. Here we ablate this choice: keeping all other settings identical, we train two Stage-3 models, one \emph{with} and one \emph{without} the pure-pinhole data, and evaluate both on OmniScene-Full (heterogeneous six-view), OmniScene-Quad (four-view fisheye), and OmniOcc (real-world pinhole). This isolates the effect of pure-pinhole replay on the heterogeneous and fisheye domains as well as the pure-pinhole domain.

\begin{table}[h]
\centering
\caption{Stage-3 pure-pinhole replay ablation. We compare Stage-3 training with and without pure-pinhole data on heterogeneous (OmniScene-Full), fisheye (OmniScene-Quad), and real-world pinhole (OmniOcc) benchmarks.}
\label{tab:appendix_pinhole_replay}
\footnotesize
\setlength{\tabcolsep}{2.0pt}
\begin{tabular*}{\textwidth}{@{\extracolsep{\fill}}llccc@{}}
\toprule
Benchmark & Stage-3 variant & Scale AbsRel $\downarrow$ & AbsRel $\downarrow$ & RMSE $\downarrow$ \\
\midrule
\multirow{2}{*}{OmniScene-Full} & w/o pinhole replay & 0.1057 & 0.1041 & 1.5028 \\
 & Ours & 0.1181 & 0.1021 & 1.5993 \\
\midrule
\multirow{2}{*}{OmniScene-Quad} & w/o pinhole replay & 0.1241 & 0.1033 & 1.6819 \\
 & Ours & 0.1268 & 0.1138 & 1.6481 \\
\midrule
\multirow{2}{*}{OmniOcc} & w/o pinhole replay & 0.2061 & 0.0883 & 0.2881 \\
 & Ours & 0.0670 & 0.0656 & 0.2584 \\
\bottomrule
\end{tabular*}
\end{table}

As shown in \cref{tab:appendix_pinhole_replay}, removing pure-pinhole data from Stage 3 markedly degrades pinhole depth on OmniOcc while leaving OmniScene-Full and OmniScene-Quad largely unchanged, confirming that the replay restores multi-view pinhole capability at negligible cost to the heterogeneous and fisheye settings.

\section{OmniOcc: Real-world Surround-view Benchmark}
\label{sec:appendix_omniocc}
To assess our model's robustness in real-world data, we curate OmniOcc, a real-world indoor dataset for multi-view geometric parsing. The hardware setup consists of four stereo camera pairs $\text{cam}_{0}$--$\text{cam}_{3}$, each providing a left and a right view for 8 synchronized RGB views in total at $1280 \times 1088$ resolution, together with a co-registered LiDAR sensor, establishing omnidirectional coverage under a shared metric coordinate frame. OmniOcc spans 25 diverse indoor scenes across varying architectural layouts (e.g., offices, hotel rooms, corridors, and elevator lobbies), with each scene stored as a synchronized keyframe snapshot.

OmniOcc is used only for evaluation and is never included in training. For the real-world multi-view depth benchmark, we use six of the eight available views: the left and right images of the first three stereo pairs $\text{cam}_0$, $\text{cam}_1$, $\text{cam}_2$, while the fourth pair $\text{cam}_3$ is excluded. Per-view metric supervision is obtained by projecting the registered LiDAR point cloud onto the corresponding camera plane and retaining only valid observed pixels after visibility and projection filtering. Therefore, we describe the supervision as metric LiDAR-projected supervision on valid observed pixels rather than dense or pixel-perfect ground truth. This protocol accounts for the inherent sparsity of LiDAR projection as well as occlusion and visibility constraints in real-world capture.

For reproducibility, we report OmniOcc statistics and preprocessing together with the benchmark results, including the total number of keyframes, the valid-pixel ratio after projection, the projection and occlusion filtering rules, and the synchronization and calibration procedure. All 25 scenes are used purely for evaluation, and all depth errors are computed only over valid LiDAR-observed pixels.

\section{Evaluation Protocol}
\label{sec:appendix_eval_detail}
\label{supp:datasplit}

\paragraph{OmniScene split.} OmniScene contains 103 scenes, 564 motion sequences, and approximately 266K synchronized six-view frames. We split it \emph{by scene} so that the training and test sets share no scene: the training set has 91 scenes (541 sequences) and the test set has 12 held-out scenes (23 sequences, $10{,}420$ frames). Following common practice, we additionally monitor training on a small validation subset of 62 sequences drawn from 44 training scenes; this subset is used only to track convergence and select checkpoints and is not a held-out set, whereas the 12 test scenes are never seen during training. All three OmniScene benchmarks---OmniScene-Full (six-view heterogeneous), OmniScene-Quad (four-view fisheye), and OmniScene-Single (monocular fisheye)---are evaluated on these 12 held-out test scenes.

\paragraph{Evaluation benchmarks.} Fisheye and heterogeneous evaluation use the held-out OmniScene test scenes (OmniScene-Full/Quad/Single) and KITTI360~\cite{liao2022kitti360}. Pinhole evaluation uses ETH3D~\cite{schops2017eth3d}, ScanNet++ v2~\cite{yeshwanth2023scannetpp}, and our real-world OmniOcc benchmark \cref{sec:appendix_omniocc}. ETH3D and OmniOcc are used solely for evaluation. ScanNet++ v2 and KITTI360 appear in both training and evaluation, and we keep their evaluation data strictly disjoint from training and validation. For ScanNet++ v2, we evaluate on a fixed two-view protocol over 10 held-out scenes (\texttt{062e5a23a6}, \texttt{13b4efaf62}, \texttt{3c8d535d49}, \texttt{6c14d5fd01}, \texttt{95b9971d01}, \texttt{9bfbc75700}, \texttt{a4c043ac48}, \texttt{cba701332a}, \texttt{e3b3b0d0c7}, \texttt{e667e09fe6}); none of these scenes appears in the training or validation set. For KITTI360, we use sequence~0 as the test set and all remaining sequences for training and validation, following the same protocol as DAC~\cite{zhang2025depthanycamera} and UniDAC~\cite{ganesan2026unidac}. All reported depth metrics are computed on these held-out evaluation sets.

\clearpage
\begin{figure*}[p]
    \centering
    \includegraphics[width=\textwidth,height=0.45\textheight,keepaspectratio]{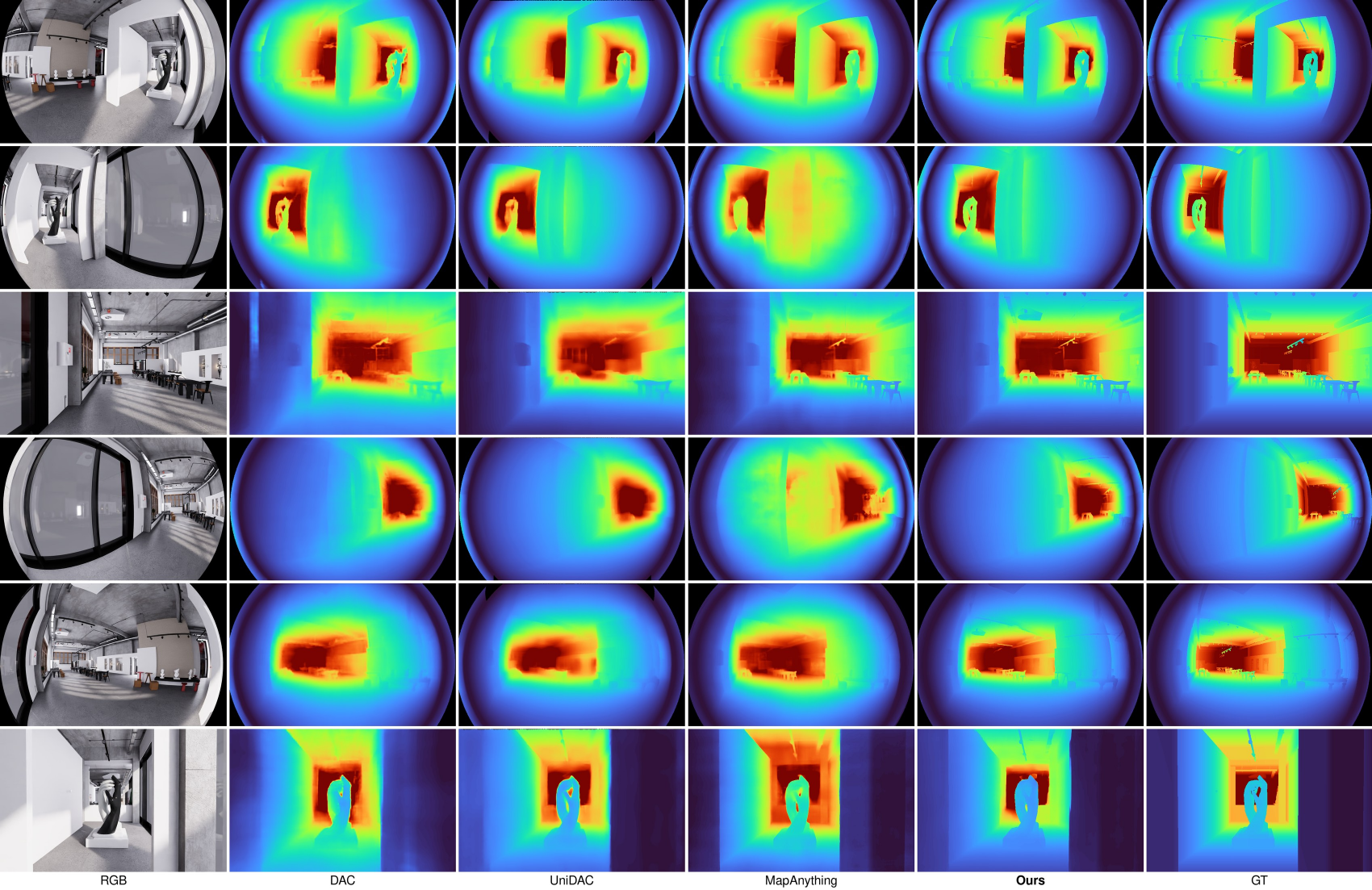}

    \vspace{0.5em}

    \includegraphics[width=\textwidth,height=0.45\textheight,keepaspectratio]{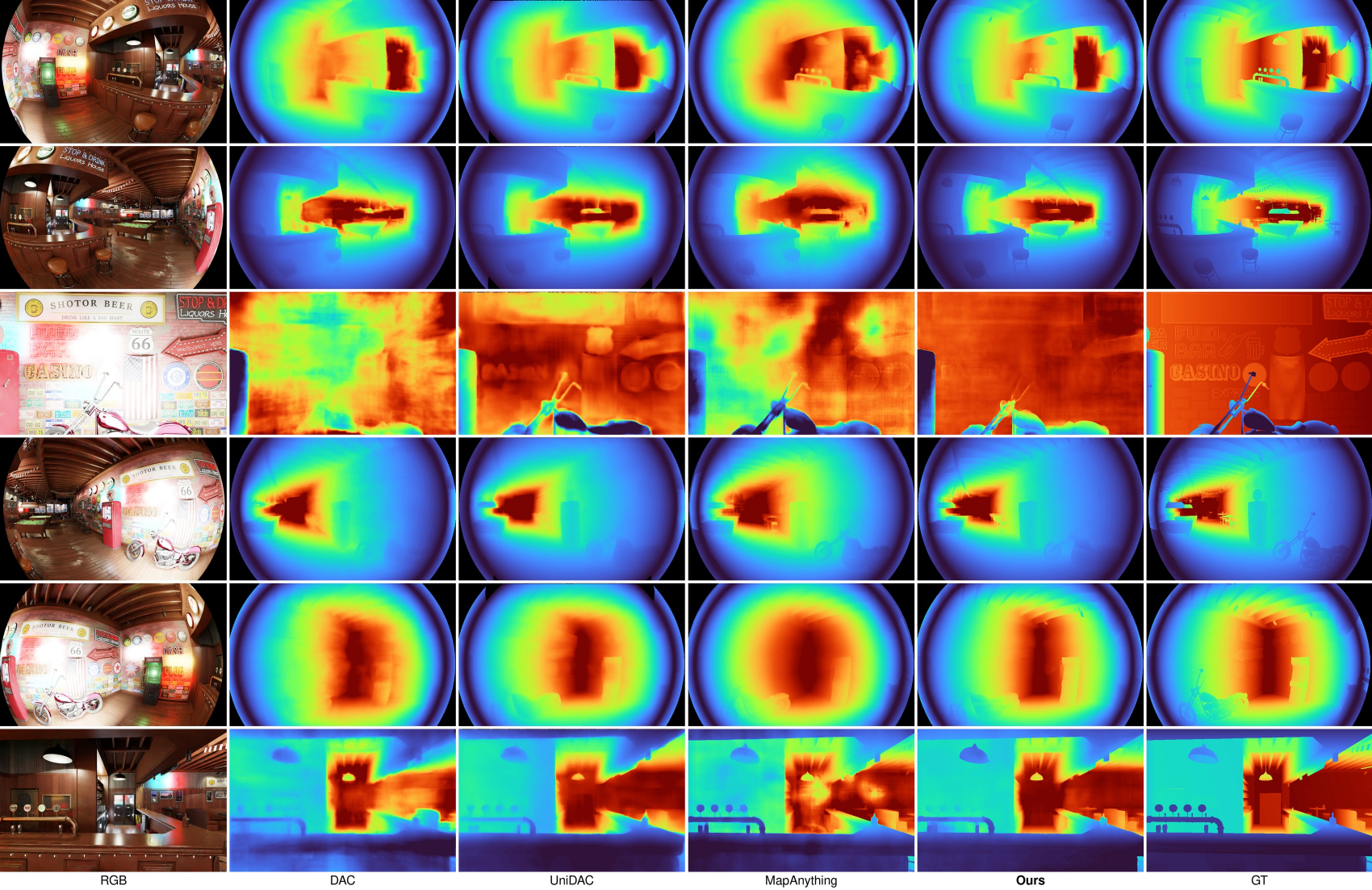}
    \caption{Qualitative comparisons on OmniScene scenes with a six-camera heterogeneous rig (two pinhole cameras and four fisheye cameras). Each example contains RGB, DAC, UniDAC, MapAnything, \ours, and GT, showing depth predictions across mixed camera types.}
    \label{fig:supply3}
    \label{fig:supply2}
\end{figure*}

\clearpage
\begin{figure*}[p]
    \centering
    \includegraphics[width=\textwidth,height=0.45\textheight,keepaspectratio]{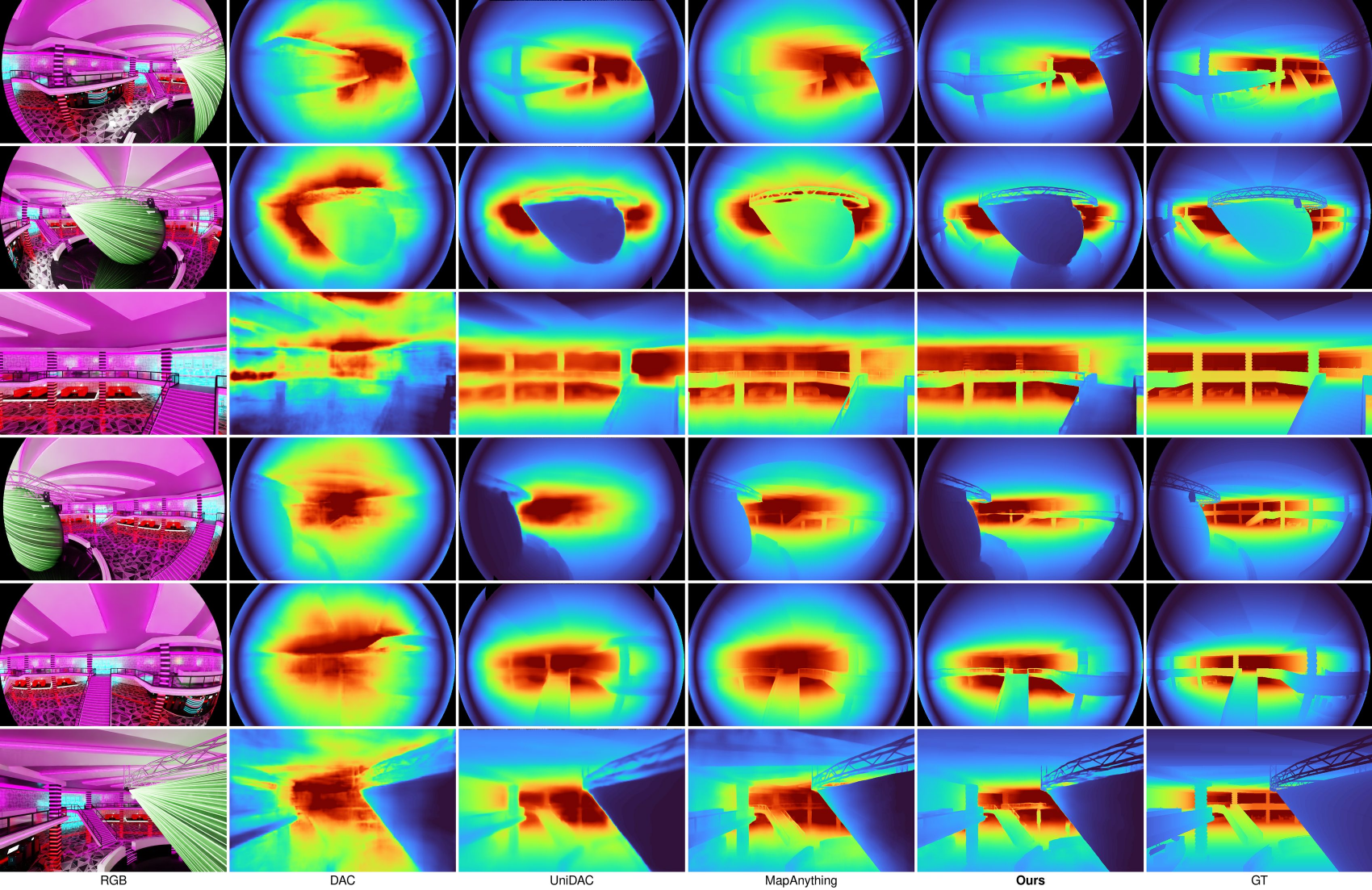}

    \vspace{0.5em}

    \includegraphics[width=\textwidth,height=0.45\textheight,keepaspectratio]{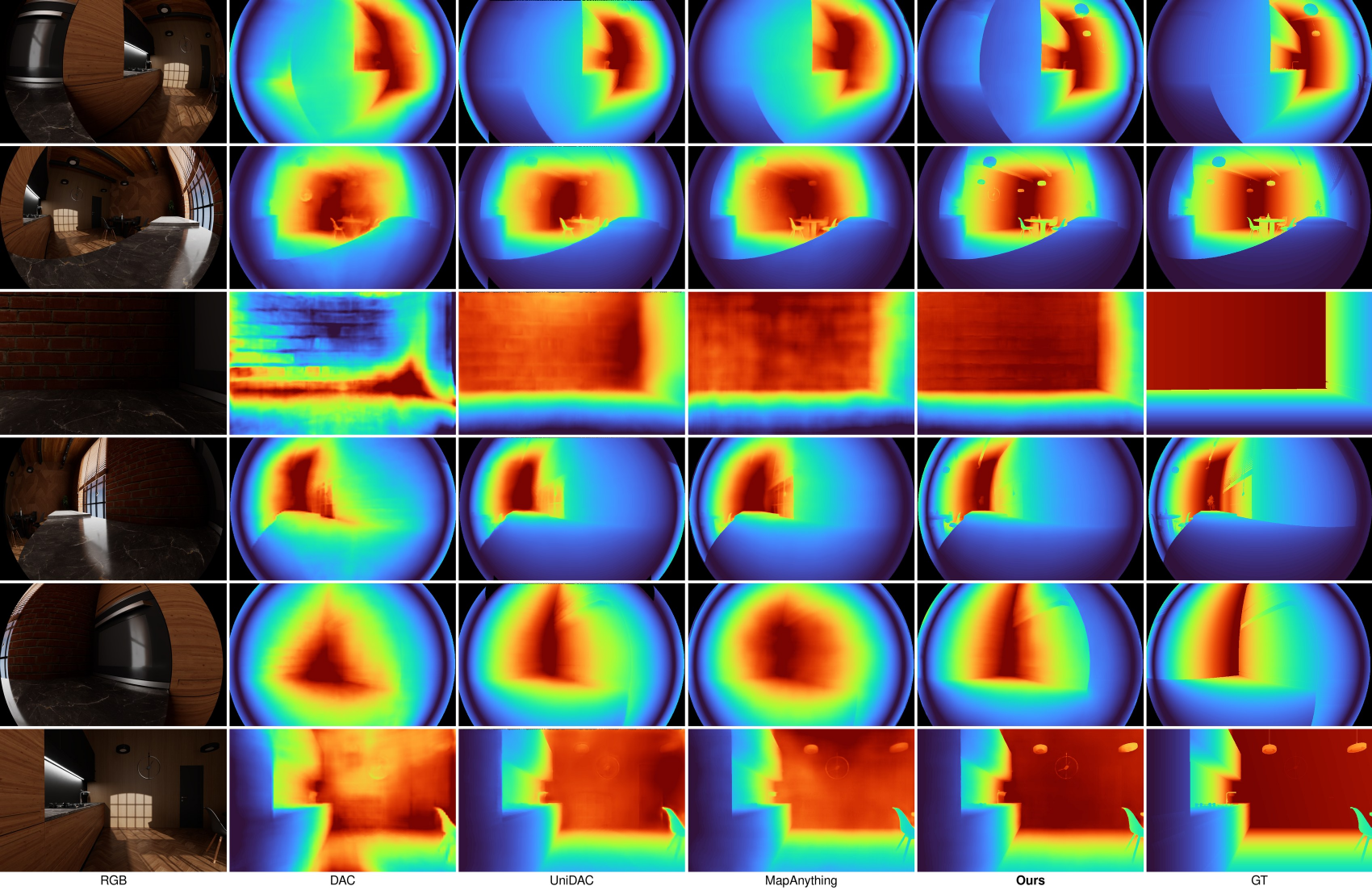}
    \caption{Additional qualitative comparisons on OmniScene scenes under the same six-camera heterogeneous setting (two pinhole cameras and four fisheye cameras). The rows compare RGB, DAC, UniDAC, MapAnything, \ours, and GT, highlighting the metric-depth consistency of \ours.}
    \label{fig:supply4}
    \label{fig:supply1}
\end{figure*}
\clearpage

\end{document}